%% file: main.tex
\documentclass[10pt,twocolumn,letterpaper]{article}

\usepackage{iccv}              
\usepackage{times}
\usepackage{epsfig}
\usepackage{graphicx}
\usepackage{amsmath}
\usepackage{amssymb}
\usepackage{graphicx}
\usepackage{amsmath}
\usepackage{amssymb}
\usepackage{booktabs}
\usepackage{amsthm}
\usepackage{algorithm}
\usepackage{algorithmic}
\usepackage[algo2e, ruled]{algorithm2e}
\usepackage{nicefrac}       
\usepackage{microtype} 
\usepackage{multirow}
\usepackage{xcolor}
\usepackage{color, colortbl}
\definecolor{Gray}{gray}{0.9}

\usepackage{babel}
\providecommand{\lemmaname}{Lemma}
\providecommand{\theoremname}{Theorem}

\theoremstyle{plain}

\theoremstyle{plain}

\newcolumntype{a}{>{\columncolor{Gray}}c}
\newcolumntype{H}{>{\setbox0=\hbox\bgroup}c<{\egroup}@{}}
\newcolumntype{Z}{>{\setbox0=\hbox\bgroup}c<{\egroup}@{\hspace*{-\tabcolsep}}}

\newcommand\csm[1]{\textcolor{black}{#1}}

\usepackage[pagebackref,breaklinks,colorlinks,allcolors=iccvblue]{hyperref}
\usepackage[capitalize]{cleveref}
\crefname{section}{Sec.}{Secs.}
\Crefname{section}{Section}{Sections}
\Crefname{table}{Table}{Tables}
\crefname{table}{Tab.}{Tabs.}

\input{preamble}

\definecolor{iccvblue}{rgb}{0.21,0.49,0.74}

\def\paperID{2} 
\def\confName{ICCV}
\def\confYear{2025}

\title{Toward a Holistic Approach to Continual Model Merging}

\author{
Hoang Phan$^1$ \quad Sungmin Cha$^1$ \quad Lam Tran$^2$ \quad Qi Lei$^1$ \\
{\tt\small \{hvp2011, sungmin.cha, ql518\}@nyu.edu}, \tt\small ttran53@ur.rochester.edu
\\
 $^1$New York University \quad \quad  $^2$University of Rochester
}

\begin{document}
\maketitle
\input{sec/0_abstract}    
\input{sec/1_intro}
\input{sec/2_related_work}

\input{sec/3_background}
\input{sec/4_method}
\input{sec/5_experiment}
\input{sec/6_conclusion}

\clearpage
\section*{Acknowledgments}
We thank anonymous reviewers for helpful feedback. QL acknowledges support of NSF DMS-2523382 and DOE Office of Science under Award \#DE-SC0024721.

{
    \small
    \bibliographystyle{ieeetr}
    \bibliography{main}
}
\appendix
\onecolumn
\input{sec/7_supp}

\end{document}

%% file: preamble.tex
\usepackage{epsfig}
\usepackage{graphicx}
\usepackage{amsmath}
\usepackage{amssymb}
\usepackage{graphicx}
\usepackage{amsmath}
\usepackage{amssymb}
\usepackage{booktabs}
\usepackage{amsthm}
\usepackage{amsfonts}
\usepackage{algorithm}
\usepackage{algorithmic}
\usepackage{nicefrac}       
\usepackage{microtype} 
\usepackage{multirow}
\usepackage{wrapfig}
\usepackage{caption}
\usepackage{subcaption}
\makeatother
\usepackage{xcolor}
\usepackage{color, colortbl}
\definecolor{Gray}{gray}{0.9}

\usepackage{babel}
\providecommand{\lemmaname}{Lemma}
\providecommand{\theoremname}{Theorem}

\newcolumntype{a}{>{\columncolor{Gray}}c}
\newcolumntype{H}{>{\setbox0=\hbox\bgroup}c<{\egroup}@{}}
\newcolumntype{Z}{>{\setbox0=\hbox\bgroup}c<{\egroup}@{\hspace*{-\tabcolsep}}}

\usepackage[capitalize]{cleveref}
\crefname{section}{Sec.}{Secs.}
\Crefname{section}{Section}{Sections}
\Crefname{table}{Table}{Tables}
\crefname{table}{Tab.}{Tabs.}

%% file: sec/0_abstract.tex
\begin{abstract}
We present a holistic framework for \textbf{C}ontinual \textbf{M}odel \textbf{M}erging (CMM) that intervenes at three critical stages: pre-merging, during merging, and post-merging—to address two fundamental challenges in continual learning. In particular, conventional approaches either maintain a growing list of per-domain task vectors, leading to scalability issues or rely solely on weight-space merging when old data is inaccessible, thereby losing crucial functional information. Our method overcomes these limitations by first fine-tuning the main model within its tangent space on domain-specific data; this linearization amplifies per-task weight disentanglement, effectively mitigating across-task interference. During merging, we leverage functional information from available optimizer states beyond mere parameter averages to avoid the need to revisit old data. Finally, a post-merging correction aligns the representation discrepancy between pre- and post-merged models, reducing bias and enhancing overall performance—all while operating under constant memory constraints without accessing historical data. Extensive experiments on standard class-incremental and domain-incremental benchmarks demonstrate that our approach not only achieves competitive performance but also provides a scalable and efficient solution to the catastrophic forgetting problem. Our code is available at \url{https://github.com/VietHoang1512/CMM}.
\end{abstract}

%% file: sec/1_intro.tex
\section{Introduction}
\label{sec:intro}

\csm{The rapid proliferation of large-scale pre-trained models has fundamentally reshaped artificial intelligence research.}
Models, such as GPT-4~\citep{achiam2023gpt, hurst2024gpt}, Llama~\citep{touvron2023llama, grattafiori2024llama}, Deepseek~\citep{liu2024deepseek, liu2024deepseek3, guo2025deepseek} and Stable Diffusion~\citep{rombach2022high} or Sora~\citep{brooks2024video} exemplify the transformative power of training extensive neural networks on diverse datasets, offering robust performance across numerous tasks and domains. 
Platforms like HuggingFace\footnote{https://huggingface.co/models} currently host approximately 1.5  million publicly available models, each fine-tuned or pretrained with unique data sources, architectures, and training methodologies. 
\csm{This growing repository presents both an opportunity and a challenge: while specialized models can excel in their respective domains, their fragmented nature raises questions about how best to leverage them for broader applications.}

\csm{Model merging has emerged as a promising direction to address this challenge~\cite{li2023deep, yang2024model}.}
Model merging involves combining the parameters of multiple pre-trained models into a single unified model, aiming to retain or even enhance the individual strengths of each constituent model. Unlike traditional ensemble methods that aggregate model outputs at inference, parameter merging directly integrates learned representations within a single model, thereby reducing inference cost and storage requirements, while potentially boosting generalization and robustness.

Beyond \csm{efficiency gains}, model merging provides versatile methodological tools applicable to several critical areas of machine learning, including multi-task learning \citep{yang2023adamerging, ahmadian2024mix, li2024improving}, federated learning \cite{li2023revisiting, jia2024dapperfl,saadati2024dimat}, and particularly continual learning \citep{chitale2023task, araujo2024learning, quarantiello2024adaptive}. 
In multi-task learning, merging specialized models trained on distinct tasks can yield a comprehensive model capable of simultaneous task proficiency. 
In federated learning scenarios, model merging facilitates privacy-preserving aggregation of locally trained models without exchanging sensitive training data. 
Most notably, in continual learning, merging allows models to sequentially incorporate new knowledge without catastrophic forgetting, significantly advancing the stability and scalability of lifelong learning systems.

\begin{figure}[!ht]
    \centering
    \vspace*{-2mm}    
     \includegraphics[width=1\columnwidth,]{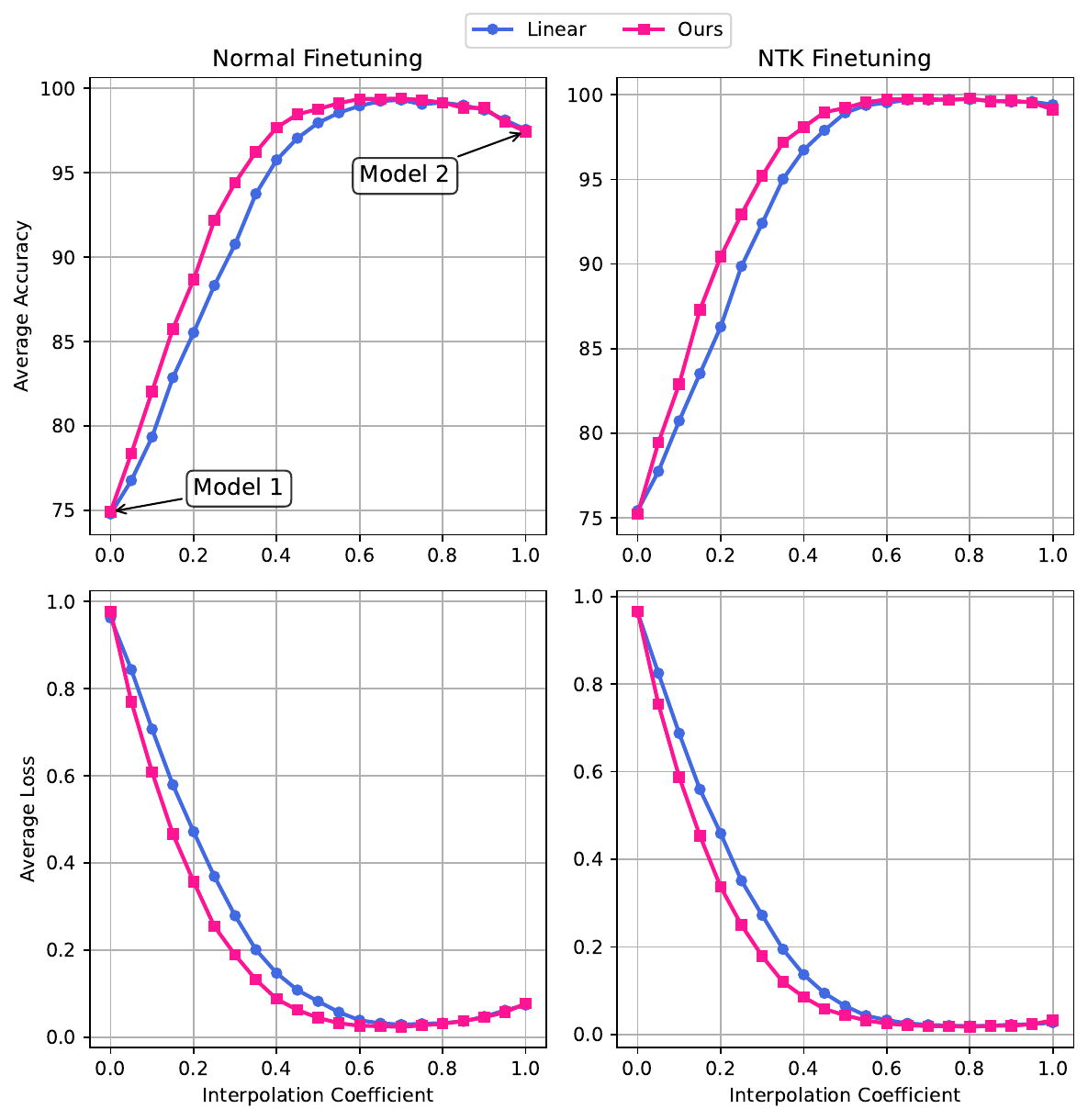}
    \vspace{-3mm}
    \caption{Average accuracy score and loss value when interpolating models after training on first and second tasks, using linear averaging and our proposed merging method.}
    \vspace*{-2mm}    
    \label{fig:merge}
\end{figure}

Despite growing interest and wide application, existing merging techniques predominantly focus on \csm{the ``during-merging'' phase,} where practitioners combine multiple models without access to their original training data \citep{goddard2024arcee, tang2024fusionbench}. 
\csm{This limitation can lead to suboptimal performance, as naive merging approaches often overlook critical functional information embedded in the training process. Additionally, the number of stored models can grow linearly with the number of tasks, posing scalability challenges.}
\csm{To overcome these limitations, we propose CMM - a holistic framework that systematically addresses model merging across three key phases, pre-merging, during-merging, and post-merging.}
By leveraging task-specific data, our method exploits valuable functional information, \csm{leading to} improved merging performance compared to \csm{conventional} linear interpolation methods. Figure \ref{fig:merge} \csm{highlights} this improvement, \csm{illustrating} accuracy and loss curves when interpolating between models trained on the first two tasks of the Cars~\citep{krause20133d} dataset, where our proposed method (pink) \csm{consistently} outperforms linear interpolation (blue) \csm{across} all settings.

\csm{Based on this finding, this paper aims to address key limitations in existing merging-based continual learning approaches.}
Specifically, we propose to first fine-tune task-specific models within the tangent space of a pre-trained model, effectively mitigating conflicts during subsequent model merging. 
Next, we derive a closed-form merging solution that naturally leverages expressive functional information \csm{without additional computational cost}. 
Furthermore, we identify and analyze the representation mismatch between merged models and previously learned representations, motivating a representation refinement step as a beneficial post-merging procedure. In summary, our key contributions are:
\begin{itemize}
    \item We shed \csm{light} on the fundamental pitfalls encountered when naively applying existing model merging techniques to continual learning scenarios, particularly emphasizing issues such as limited scalability and neglect of available task-specific functional information.
    \item Based on these insights, we propose a unified merging-based framework specifically designed to alleviate catastrophic forgetting not only \csm{during the merging phase but also before and after merging}.
    \item We conduct extensive experiments across several popular benchmarks in both class-incremental and domain-incremental learning settings, empirically validating the effectiveness and versatility of our proposed approach.
\end{itemize}

%% file: sec/2_related_work.tex
\section{Related work}
\label{sec:related}
\subsection{Continual learning}
Continual learning (CL) aims to develop models capable of adapting sequentially to new tasks while maintaining performance on previously learned tasks. The main challenge in CL is catastrophic forgetting, where acquiring new information interferes with existing knowledge~\citep{kirkpatrick2017overcoming,rebuffi2017icarl}. Existing approaches to mitigate forgetting can broadly be categorized into three groups: regularization-based, rehearsal-based, and architecture-based methods. Regularization-based approaches constrain updates to parameters crucial for previous tasks, thus reducing interference~\citep{li2017learning,kirkpatrick2017overcoming}. Architectural methods expand or adapt the model architecture dynamically, assigning distinct parameter subsets to individual tasks to minimize conflict~\citep{rusu2016progressive,yan2021dynamically}. Rehearsal-based methods retain exemplars from previous tasks in a memory buffer, leveraging them during training of subsequent tasks to prevent forgetting~\citep{rebuffi2017icarl,buzzega2020dark}. However, rehearsal-based methods raise concerns about data privacy and storage requirements. \csm{Given these limitations}, recent efforts have increasingly focused on memory-free continual learning, utilizing techniques such as generative replay~\citep{shin2017continual}, prompt-based adaptation~\citep{wang2022learning,wang2022dualprompt}, and leveraging pre-trained models to achieve robust performance without data storage~\citep{wang2023comprehensive,smith2023coda}.

\subsection{Model merging}
Model merging \csm{aims} to combine multiple pre-trained or task-specific models into a unified model that integrates their strengths. Early merging methods typically average model parameters directly, based on the mode connectivity assumption that different parameter solutions lie in similar regions of the loss landscape~\citep{wortsman2022model,ilharco2023task}. Subsequent approaches introduced more sophisticated alignment-based strategies, such as weight matching and permutation-invariant transformations, to better handle parameter disparities between merged models~\citep{matena2022merging,jin2023dataless,ainsworth2022git}. Recent techniques, including Task Arithmetic~\citep{ilharco2023task}, TIES-Merging~\citep{yadav2023ties}, and Fisher merging~\citep{matena2022merging,dhawan2024leveraging}, further refine merging by considering task-specific importance or parameter influence, significantly improving generalization. Despite these advances, most existing merging methods assume simultaneous access to multiple models, which restricts their direct applicability to continual learning scenarios. For instance, MagMax~\citep{marczak2024magmax} retains one task vector per task, resulting in linear memory growth with respect to the number of tasks. It aggregates these vectors by selecting, for each parameter, the value with the largest absolute magnitude across tasks. Since MagMax performs arithmetic operations purely in the parameter space, this method neglects valuable information encoded in task-specific data. \csm{The most} closely related to our work is CoFiMA~\citep{marouf2024weightedensemblemodelsstrong}, \csm{which extends Fisher Merging to continual learning by applying an online variant to merge models sequentially.}
In contrast, our method directly approximates the Fisher Information Matrix by leveraging the second-moment statistics readily available from standard deep-learning optimizers such as Adam~\citep{kingma2014adam} or AdamW~\citep{loshchilov2018decoupled}. 
\csm{Furthermore, our method mitigates task interference during the merging process and incorporates feature refinement post-merging, enabling effective sequential model merging while preserving representational consistency across tasks.}

%% file: sec/3_background.tex
\section{Background}
\label{sec:background}
\subsection{Continual learning}
\csm{Continual learning for large pre-trained models is crucial for enabling efficient knowledge integration across multiple tasks. In this regard, we focus on continual fine-tuning, where a pre-trained model $\theta$ is given the goal is to incrementally learn knowledge from 
\(N\) different domains or tasks, denoted as \(\{\mathcal{D}_{1}, \mathcal{D}_{2}, \dots, \mathcal{D}_{N}\}\).
Each domain or task's dataset \(\mathcal{D}_{n}=\{\boldsymbol{X}_n,\boldsymbol{Y}_n\}\) is accessible only during the \(n\)-th fine-tuning phase. We investigate two key challenging continual learning scenarios: Class-Incremental Learning (CIL) and Domain-Incremental Learning (DIL).}
In CIL, the class labels \(y^j_{n} \in \mathcal{Y}_{n}\) from  the incoming domain \(\mathcal{D}_{n}\) are disjoint from the set of seen class labels $\mathcal{Y}_{1:n-1}$. 
\csm{In contrast, DIL assumes that all domains share the same label space $\mathcal{Y}$.}
For both scenarios, during inference at any task, \csm{models must} classify input images into all seen classes so far without knowing their domain \csm{identity information}.

In this paper, we primarily focus on \textit{data-free continual learning}, \csm{where data from prior domains is inaccessible}. 
\csm{In this setting, the only information retained from past tasks during training on $t$-th domain is the optimally fine-tuned model parameters from the previous task $\theta_{t-1}^*$.}

\subsection{CLIP and nearest mean classifier}

\csm{CLIP \citep{radford2021learning} is a dual-encoder model that learns a joint embedding space for images and text. It consists of an image encoder \( f(\cdot) \) and a text encoder \( g(\cdot) \), mapping an image \( I \) and text prompt \( T \) into normalized embeddings \( z_I = f(I) \) and \( z_T = g(T) \), respectively. Image classification using CLIP can be performed by computing the cosine similarity between an image embedding and class text embeddings: $s_k = \frac{z_I \cdot z_{T_k}}{\|z_I\|\|z_{T_k}\|}$.}
The similarity scores are transformed into probabilities using a softmax function with temperature \( \tau \):
\[
p(y=k \mid I) = \frac{\exp(s_k/\tau)}{\sum_{j=1}^K \exp(s_j/\tau)}
\]
This formulation allows CLIP to perform zero-shot classification by leveraging the learned joint embedding space without the need for task-specific training. In all our experiments, the encoder \( g(\cdot) \) is frozen and we use the nearest mean classifier to classify input images using mean-class embeddings.

\subsection{Adam and Fisher Information Matrix}
 \csm{Adam~\citep{kingma2014adam} is an adaptive gradient-based optimizer that utilizes first- and second-moment estimates for efficient optimization. AdamW~\citep{loshchilov2017decoupled} further improves it by decoupling weight decay from gradient updates. The full update rule with decay parameter $\lambda$ is given by:}
\begin{align} 
&\boldsymbol{g}_t \leftarrow \nabla f_t\left(\boldsymbol{\theta}_{t-1}\right)+\lambda \boldsymbol{\theta}_{t-1}\nonumber \\
& \boldsymbol{m}_t \leftarrow \beta_1 \boldsymbol{m}_{t-1}+\left(1-\beta_1\right) \boldsymbol{g}_t \quad \text{// \csm{first-moment estimate}}\nonumber\\ 
& \boldsymbol{v}_t \leftarrow \beta_2 \boldsymbol{v}_{t-1}+\left(1-\beta_2\right) \boldsymbol{g}_t^2 \quad \text{// \csm{second-moment estimate}} \nonumber\\ 
& \widehat{\boldsymbol{m}_t} \leftarrow \boldsymbol{m}_t /\left(1-\beta_1^t\right) \nonumber\\ 
& \widehat{\boldsymbol{v}_t} \leftarrow \boldsymbol{v}_t /\left(1-\beta_2^t\right) \nonumber\\
& \theta_t \leftarrow \theta_t-\gamma \widehat{\boldsymbol{m}_t} /\left(\sqrt{\widehat{\boldsymbol{v}_t}}+\epsilon\right) \label{eq:adam}
\end{align}
where $t$ denotes the time step, $\boldsymbol{g}_t \in \mathbb{R}^n$ the gradient vector and $\eta>0$ the step size. 
The exponential moving average (EMA) time scales of the first gradient moment $\boldsymbol{m}_t$ and second moment $\boldsymbol{v}_t$ are set by $0 \leqslant \beta_1, \beta_2<1$. 
\csm{Note that, when $\beta_1 = 0$, $\boldsymbol{m}_t$ corresponds to the diagonal Fisher Information matrix at the $t$-th iteration, which is widely used for preconditioning in various optimizers~\citep{amari1998natural,eschenhagen2023kronecker,hwang2024fadam}}

\subsection{Neural tangent kernel} 



In the vicinity of the initial weights $\theta_0$, a neural network's behavior can be closely approximated by the following first-order Taylor expansion: 
\begin{equation} f(x;\theta) \approx f(x;\theta_0) + (\theta-\theta_0)^\top \nabla_{\theta} f(x;\theta_0).\label{eq:linearization} \end{equation} 
\csm{This approximation corresponds to a kernel-based predictor using the \textit{Neural Tangent Kernel} (NTK)}, represented as $k_{\text{NTK}}(x, x') = \nabla_{\theta} f(x;\theta_0)^\top \nabla_{\theta} f(x';\theta_0)$. It establishes a neural tangent space wherein the connection between weights and functional outputs is linear. 
\csm{Note that, as network width approaches infinity, this linearization remains valid throughout training~\citep{jacot2018neural, arora2019exact, lee2019wide}.}

Nonetheless, this linear model often does not hold at finite network widths, as it fails to adequately represent the evolution of parameters during training, according to Equation \eqref{eq:linearization}. In these instances, the training dynamics operate in a \textit{non-linear regime}. Conversely, in scenarios such as fine-tuning, where the evolution of parameters in many pre-trained models is relatively minor, the training largely remains within the tangent space, thus the linear approximation in Equation \eqref{eq:linearization} effectively mirrors the network's behavior \cite{malladi2022kernel,ortizjimenez2021linear,zancato2020predicting,deshpande2021linearized,yuce2022inrs}, indicative of a \textit{linear regime}.

%% file: sec/4_method.tex
\section{Proposed method}
\label{sec:method}
To solve the above problems, we propose a new algorithm.
\csm{In this section, we introduce a novel algorithm to address the challenges associated with model merging in continual learning. Our framework improves upon traditional methods by incorporating second-order information, enhancing task-specific representation alignment, and reducing task interference during the merging process.}

\subsection{\csm{Merging during task progression}}
\csm{Let $\mathcal{D}_i$ denote a sequence of tasks indexed by $i = 1, ..., N$.}
For each task \(i\), the model with parameters \(\theta_i\) is fine-tuned until convergence. 
The loss function for $i$-th task is defined as:
\[
L_i(\theta)
\;=\;
\mathbb{E}_{(x,y)\sim \mathcal{D}_i}
\Big[\,\ell\big(f_\theta(\boldsymbol{X}),\,\boldsymbol{Y}\big)\Big],
\]
\csm{where $l$ represents the loss between the model prediction $f_\theta(\boldsymbol{X})$ and the true label $\boldsymbol{Y}$.}
Since \(\theta_i\) is fine-tuned on task \(i\), it implies that \(\theta_i\) is (approximately) a local minimum of \(L_i(\theta)\).  A standard second-order Taylor expansion of \(L_i(\theta^*)\) with $\theta^*$ around \(\theta_i\) is \csm{given by}:
\begin{align*}
  L_i(\theta^*) 
~\approx~
& \;L_i(\theta_i)
+
\nabla_\theta L_i(\theta_i)^\top\!\,\bigl(\theta^* - \theta_i\bigr)
\;\nonumber\\
& + \tfrac{1}{2}\,\bigl(\theta^* - \theta_i\bigr)^\top\,\nabla^2_\theta L_i(\theta_i)\,\bigl(\theta^* - \theta_i\bigr).  
\end{align*}
\csm{Since} \(\theta_i\) is a local optimum for the \(i\)-th task, \csm{we have} $\nabla_\theta L_i(\theta_i) \approx 0$,
which eliminates the first-order term. Therefore, the expansion simplifies to:
\begin{align}
L_i(\theta^*) 
\approx & \; L_i(\theta_i)  + \tfrac{1}{2}\,\bigl(\theta^* - \theta_i\bigr)^\top\,\nabla^2_\theta L_i(\theta_i)\,\bigl(\theta^* - \theta_i\bigr). \label{eq:taylor}
\end{align}
\csm{In the context of model merging, where task-specific data is unavailable, we approximate the Hessian matrix \(\nabla^2_\theta L_i(\theta_i)\) using the identity matrix $I$, yielding the approximation:}
\[
L_i(\theta^*)
~\approx~
L_i(\theta_i) 
\;+\; 
\tfrac{1}{2}\,(\theta^* - \theta_i)^\top I\,(\theta^* - \theta_i).
\]
\csm{Thus, the average loss across tasks is computed as:}
\[
\mathcal{L}(\theta^*)
= \frac{1}{N}\sum_{i=1}^N
L_i(\theta_i)
\;+\;
\frac{1}{2N}\,\sum_{i=1}^N 
\|\theta^* - \theta_i\|^2.
\]
Minimizing the above objective \csm{leads to the closed-form solution: }$\theta^*~=~\frac{1}{N}\sum_{i=1}^N \theta_i$, which \csm{corresponds to a simple weight averaging approach (\textit{i.e.} weight averaging).} 

However, this approach neglects crucial functional information during the merging process,  \csm{relying solely on weight space}. To address this limitation, we propose leveraging second-order information from the optimizer to more accurately approximate the Hessian matrix in Equation \ref{eq:taylor}. Without loss of generality, we assume that $N=2$ (\csm{representing the previous and current tasks}).
We thus minimize the weighted sum of per-task objectives $(1-\lambda) L_1(\theta^*)+\lambda L_2(\theta^*)$ for some $\lambda \in (0,1)$. Using the above quadratic expansions of $L_1(\theta^*)$ and $L_2(\theta^*)$ and ignoring constant terms, the objective becomes: 
$$ (1-\lambda)\left(\theta^*-\theta_1\right)^{\top} F_1\left(\theta^*-\theta_1\right)+\lambda\left(\theta^*-\theta_2\right)^{\top} F_2\left(\theta^*-\theta_2\right),$$
 where $F_i \approx \nabla_\theta^2 L_i\left(\theta_i\right)$ is \csm{approximated by} the Fisher Information Matrix (FIM).

 Taking the gradient of the above objective with respect to $\theta$ and setting it to zero yields the optimal \csm{solution}: 
\begin{equation}
    \theta^*=\left((1-\lambda) F_1+\lambda F_2\right)^{-1}\left[(1-\lambda) F_1 \theta_1+\lambda F_2 \theta_2\right] \label{eq:fisher-merge}
\end{equation}
\csm{While directly computing the FIM over the entire training set is computationally expensive, we can approximate it using second-moment information from commonly used optimizers, such as Adam~\citep{kingma2014adam}, which tracks the moving average of squared gradients $\mathbf{v}_t$ (Equation \ref{eq:adam}) as an approximation to the diagonal of the Fisher information matrix \citep{kunstner2019limitations}. Note that this approximation is efficient, requiring no gradient computation or additional data, and thus significantly reduces the computational cost compared to traditional FIM-based methods.} 

\begin{figure}[!ht]
    \centering
    \vspace*{-2mm}    
     \includegraphics[width=1\columnwidth,]{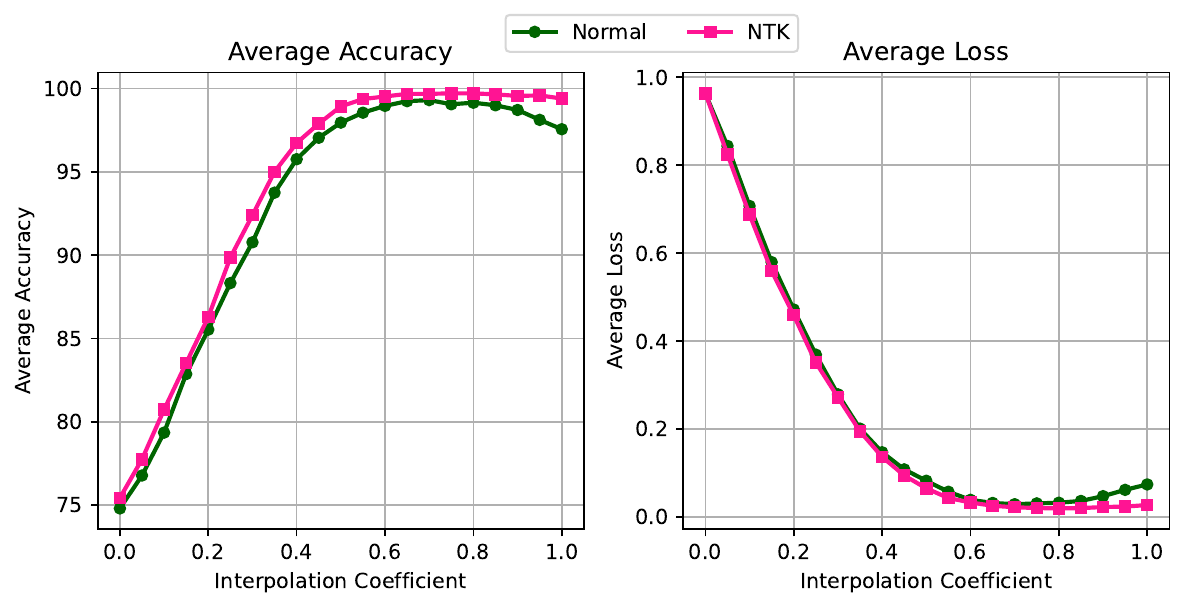}
    \vspace{-3mm}
    \caption{Average accuracy score and loss value along the linear path between models after training on  first and second tasks using normal training and our proposed linear fine-tuning method.}
    \label{fig:normal-ntk-linear}
    \vspace*{-2mm}    
\end{figure}

\subsection{Pre-merging and post-merging \csm{strategies}}
\csm{To further improve the merging process, we introduce a linear fine-tuning method aimed at better disentangling the task-specific weights and mitigating task interference \citep{ortiz2023task}.} To be more specific, Ortiz-Jimenez et al reveal that weight disentanglement is crucial for model merging, particularly to avoid negative transfer when adding linear combinations of task vectors. Motivated by this, we propose to fine-tune each model on the tangent space of the pre-trained model $\theta_0$ to effectively facilitate model merging in subsequent stages. As a simple illustrative example, Figure \ref{fig:normal-ntk-linear} shows the effectiveness of neural tangent kernel (NTK) compared with normal fine-tuning along the linear interpolation between different models.

\begin{figure}[!ht]
    \centering
    \vspace*{-2mm}
     \includegraphics[width=1\columnwidth,]{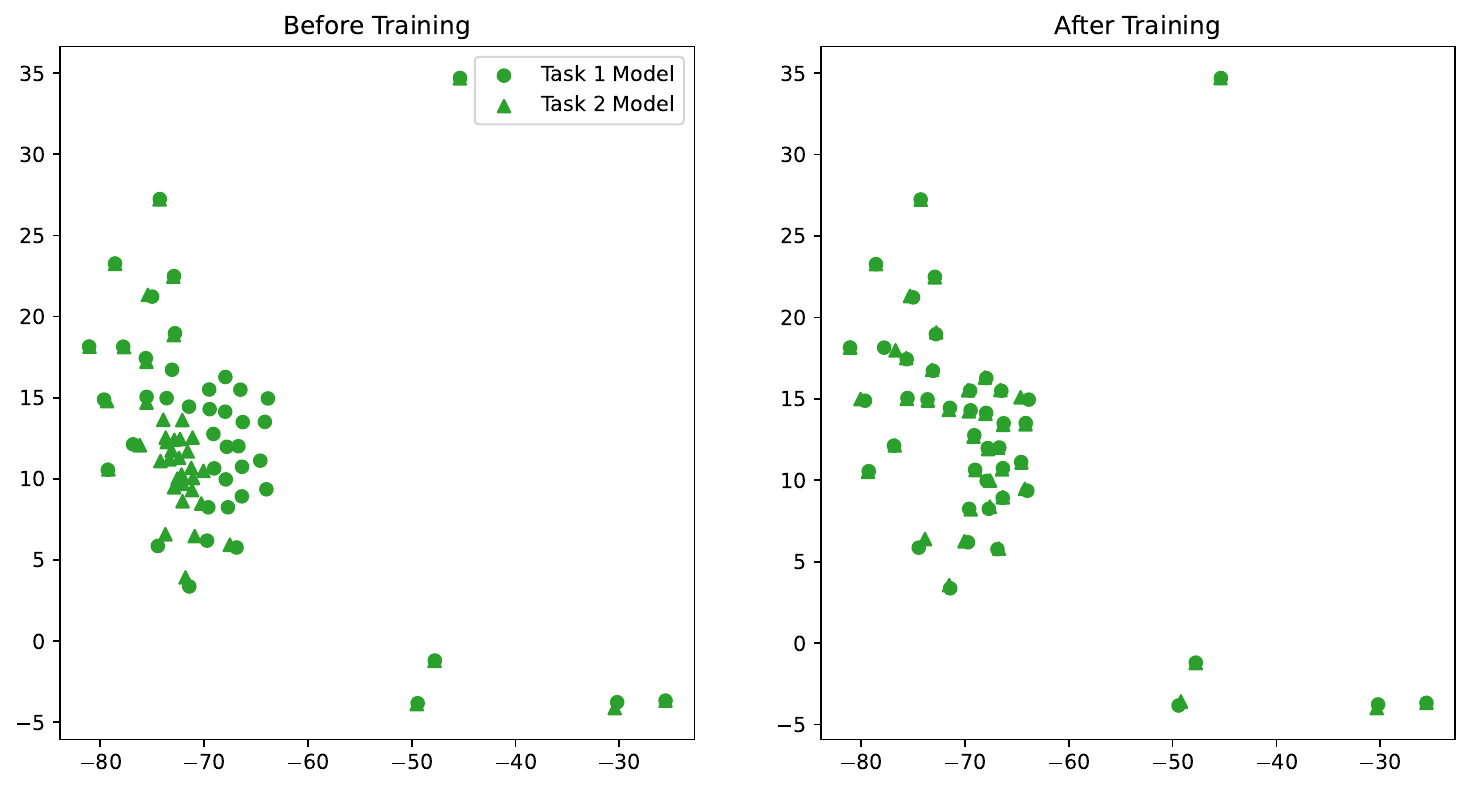}
    \vspace*{-3mm}
    \caption{Mismatch between the feature representations of models before merging and after merging (left) or between models before merging and after merging + refinement.\label{fig:mismatch}}
    \vspace*{-2mm}    
\end{figure}

\csm{Additionally, our method involves a representation alignment step post-merging to address feature representation mismatch. Figure \ref{fig:mismatch} shows the mismatch between feature representations before and after merging, which is mitigated by applying refinement after the merging process. By aligning the feature representations, we ensure that the model’s output is consistent across tasks, minimizing any detrimental effects caused by the merging process.}

\csm{Our proposed approach can be summarized in Algorithm \ref{alg:method}, which details the three-stage process of holistic continual model merging. The stages include linear fine-tuning, model merging, and representation alignment, each of which is designed to address specific limitations of conventional methods and improve task transfer in continual learning.}

\begin{algorithm}[!ht]
\caption{\label{alg:method} Holistic Continual Model Merging}
\KwIn{Pre-trained model \(\theta_0\), sequence of tasks \(\{D_t\}_{t=1}^N\), merging weight \(\lambda\)}
Initialize merged model: \(\theta_0^* \leftarrow \theta_0\)\;

\For{\(t=1,\dots,N\)}{
    \tcp{Step 1: Linear Fine-Tuning}
    Update task vector \(\tau_t\) using data \(D_t\) by solving:
    \[
    \tau_t \leftarrow \arg\min_{\tau}\; \mathcal{L}_t\Big(f_{\operatorname{lin}}(\boldsymbol{x};\theta_0+\tau), D_t\Big)
    \]
    where
    \[
    f_{\operatorname{lin}}(\boldsymbol{x};\theta_0+\tau)=f(\boldsymbol{x};\theta_0)+\tau^\top \nabla_{\theta}f(\boldsymbol{x};\theta_0)
    \]
    and update:
    \[
    \theta_t \leftarrow \theta_0+\tau_t.
    \]

    Estimate the diagonal Fisher matrix by the second moment term \(\mathbf{v}_t\) from the AdamW optimizer

\uIf{\(t > 1\)}{   

    \tcp{Step 2: Model Merging}

Merge the models as:
    \[
    \theta_t^* \leftarrow \frac{\lambda\, \mathbf{v}_t \odot \theta_t + (1-\lambda)\, \mathbf{v}_{t-1} \odot \theta_{t-1}^*}{\lambda\, \mathbf{v}_t + (1-\lambda)\, \mathbf{v}_{t-1}},
    \]
    where \(\odot\) denotes element-wise multiplication.
    
    \tcp{Step 3: Representation Alignment}
    Compute feature representations on the entire dataset \(\mathcal{D}_t\):
    \[
    \mathcal{Z}_t^{t} \leftarrow \text{FeatureExtract}(\theta_t^*, \mathcal{D}_t), \]
    \[ \mathcal{Z}_t^{t-1} \leftarrow \text{FeatureExtract}(\theta_{t-1}^*, \mathcal{D}_t).
    \]
    Calculate the representation bias:
    \[
    d_t = \frac{1}{\lvert \mathcal{D}_t\rvert}\left\|\mathcal{Z}_t^{t} - \mathcal{Z}_t^{t-1}\right\|.
    \]
    Optimize \(\theta_t^*\) (e.g. updating the last layer) to minimize \(d_t\).

    Set  \(\mathbf{v}_{t} \leftarrow {\lambda\, \mathbf{v}_t + (1-\lambda)\, \mathbf{v}_{t-1}}\) for the next iteration.
    }
}
\KwOut{Final merged model \(\theta_N^*\)}
\end{algorithm}
\vspace*{-2mm}

%% file: sec/5_experiment.tex
\section{Experimental results}
\label{sec:experiment}
To demonstrate the effectiveness of our proposed method, we conduct extensive experiments to compare our approach
with baselines on both domain-incremental and class-incremental rehearsal-free scenarios. We implemented our codebase and reproduced relevant results based on implementations of \citep{marczak2024magmax, jha2025clap4clip}.

\begin{table*}[t]{
\centering
\caption{
Comparison of continual learning methods across class-incremental setup. We report classification accuracy (\%) after the final task. The best results are in \textbf{bold} and the second best \underline{underlined}. $\Theta(1)$ and 
$\Theta(N)$ indicate constant and linear memory usage, irrespective of the number of tasks $N$.
}
\resizebox{\textwidth}{!}{
\begin{tabular}{c|l|ccc|ccc|ccc|ccc|c}
\toprule
 & & \multicolumn{3}{c|}{CIFAR100} & \multicolumn{3}{c|}{ImageNet-R} & \multicolumn{3}{c|}{CUB200} & \multicolumn{3}{c|}{Cars} & \multicolumn{1}{c}{Avg}\\
Memory & Method  & /10 & /20 & /50 & /5 & /10 & /20  & /5 & /10 & /20 & /5 & /10 & /20 & \\
\midrule

$\Theta(1)$ & Zero-shot & \multicolumn{3}{c|}{66.91} & \multicolumn{3}{c|}{77.73} & \multicolumn{3}{c|}{56.08} & \multicolumn{3}{c|}{64.71} & 67.21 \\
N/A & Joint     & \multicolumn{3}{c|}{90.94} & \multicolumn{3}{c|}{87.55} & \multicolumn{3}{c|}{81.57} & \multicolumn{3}{c|}{88.21} & 87.38 \\
\midrule
$\Theta(N)$ & RandMix \citep{marczak2024magmax}  & 77.04 & 75.36 & 72.91 & \underline{83.10} & 81.88 & 80.18  & 59.86 & 58.53 & {58.08} & 67.32 & 65.62 & \underline{64.95} & 70.40 \\
$\Theta(N)$ & MaxAbs \citep{marczak2024magmax}   & 76.75 & 74.39 & 73.04 & 83.03 & 82.33 & 80.92  & 60.15 & 58.01 & 56.59 & 67.36 & 63.55 & 58.95 & 69.59 \\
$\Theta(N)$ &  Avg  \citep{wortsman2022model}      & 77.04 & 75.29 & 72.92 & 83.08 & 81.87 & 80.27  & 59.77 & 58.44 & 58.01 & 67.37 & 65.59 & 64.88 & 70.38 \\
$\Theta(N)$ & TIES   \citep{ties}  & \underline{77.23} & 74.66 & \underline{73.76} & 83.08 & 82.27 & 80.83 & 60.94 & 58.22 & 56.97 & 70.45 & 64.90 & 61.17 & 70.37 \\
\midrule
$\Theta(1)$ & LwF  \citep{li2017learning}      & 73.45 & 72.05 & 68.84 & 81.15 & \textbf{82.97} & \underline{81.82}  & \textbf{65.12} & \underline{60.67} & \textbf{58.90} & \underline{71.72} & \textbf{69.84} & 62.98 & \underline{70.79} \\
$\Theta(1)$ & EWC  \citep{kirkpatrick2017overcoming}     & 76.24 & \underline{75.39} & 72.97 & 82.15 & 82.42 & 81.48  & 59.10 & 54.49 & 53.31 & 69.46 & 60.78 & 57.42 & 68.77 \\
$\Theta(1)$ & CMM & \textbf{77.72} & \textbf{75.95} & \textbf{73.88} & \textbf{83.41} & \underline{82.69}  & \textbf{82.00} & \underline{64.67}   & \textbf{61.13} & \underline{58.70} & \textbf{73.52}  & \underline{69.70} & \textbf{67.18} & \textbf{72.54} \\ 
\bottomrule
\end{tabular}
}
\label{tab:results-CIL}}
\vspace*{-2mm}
\end{table*}

\subsection{Evaluation Benchmarks}
\noindent{\textbf{Datasets.}} \ \ We conduct comprehensive experiments on several well-established benchmarks commonly used in continual learning research. Specifically, we use CIFAR-100 \citep{krizhevsky2009learning} and ImageNet-R \citep{hendrycks2021many} as general image recognition benchmarks and CUB-200 \citep{wah2011caltech} and Cars \citep{krause20133d} as fine-grained classification datasets. 
\csm{The CIFAR-100 dataset is split into 10/20/50 tasks, with 10/5/2 distinct classes per task, respectively.}
ImageNet-R, comprising 200 classes, is split into 5, 10, and 20 disjoint tasks to thoroughly evaluate performance across varying degrees of task granularity. 
CUB-200 and Cars datasets, representing specialized fine-grained classification scenarios, are evenly partitioned into 5/10/20 tasks, each containing an equal number of classes. 
\csm{Additionally, we evaluate our method on domain-incremental benchmarks: }Office-31 \citep{10.5555/1888089.1888106}, Office Home \citep{venkateswara2017deep}, DomainNet \citep{peng2018moment} and ImageNet-R \citep{hendrycks2021many}.

\textbf{Baselines.} We compare our proposed method against various established continual learning approaches, including general methods such as Learning without Forgetting (LwF) \citep{li2017learning} and Elastic Weight Consolidation (EWC) \citep{kirkpatrick2017overcoming}, recent model merging techniques like  Model Soup (Avg) \citep{wortsman2022model}, RandMix \citep{marczak2024magmax}, and TIES-Merging (TIES) \citep{yadav2023ties}, as well as prompt-based continual learning methods such as L2P \citep{wang2022learning}, DualPrompt \citep{wang2022dualprompt}, AttriCLIP \citep{wang2023attriclip}, and CODA-Prompt \citep{smith2023coda}. 
Additionally, we evaluate our approach against a baseline model that performs linear interpolation and examine performance against the joint-training upper bound. This comprehensive set of baselines ensures a robust and fair evaluation of our merging strategy within diverse continual learning scenarios.

\noindent{\textbf{Implementation details.}} \ \ We utilize a ViT/B16 \citep{dosovitskiy2020image} image encoder from CLIP \citep{radford2021learning} as our base model. The fine-tuning procedure follows the training protocol from previous work \citep{ilharco2022patching,ilharco2023task}, using a batch size of $128$, an initial learning rate of $1e-5$, and the AdamW \citep{loshchilov2018decoupled}  optimizer with weight decay of $0.1$ and exponential decay rates $\beta_1 = 0.9$, $\beta_2 = 0.999$. A cosine annealing learning rate schedule is applied throughout training. Images are resized to 224×224 pixels \citep{ilharco_gabriel_2021_5143773}. For CIFAR-100 and ImageNet-R, Office-Home, Office-31, DomainNet, datasets, we train each task for 10 epochs, while for the more challenging Cars and CUB-200 datasets, training extends to 30 epochs per task. To preserve the open-vocabulary capability and performance consistency, the classification layer remains frozen as per \citep{ilharco2022patching,ilharco2023task}, ensuring the model effectively retains its pre-trained knowledge base while fine-tuning.

\noindent{\textbf{Evaluation Metrics}} \ \ Following the conventional continual learning framework, we assess the overall performance of a method by reporting its average accuracy after training on all tasks. Let \( R_{N,i} \) represent the classification accuracy of task \( T_i \) after training on task \( T_N \), the average accuracy \( A_N \) is then defined as: $A_N = \frac{1}{N} \sum_{i=1}^N R_{N,i}$.

\subsection{Class-incremental experimental results} 
\csm{We evaluate our proposed method in a class-incremental setting, comparing it to various baselines across several datasets.}
Table \ref{tab:results-CIL} showcases the task-agnostic accuracy achieved by each method after the completion of the final task across multiple datasets. The \csm{``Zero-shot''} method serves as a naive baseline, while the ``Join'' training method represents an upper bound on performance, assuming \csm{full} data availability during training. 
\csm{We also distinguish between methods with constant memory usage, denoted as $\Theta(1)$, and methods with memory usage that scales linearly with the number of tasks, denoted as $\Theta(N)$.}

 While EWC \cite{hu2019overcoming} improves over the naive baselines, its performance often lags behind more recent techniques. 
 \csm{Note that our method consistently outperforms other constant-memory approaches and frequently matches or even surpasses the performance of methods with linear memory usage. For example, it achieves an average accuracy of 72.54\%, outperforming RandMix, which reaches 70.4\%.}
 Also, CMM achieves the best or second-best performance across all configurations, getting a $\approx 5\%$ margin over Zero-shot CLIP. 
 \csm{On} CIFAR100, our method achieves strong results across all incremental splits, confirming its robustness as the number of tasks grows. \csm{On} ImageNet-R, a dataset designed for studying adversarial robustness and transfer learning, our method continues to maintain a leading position among non linear-memory approaches, showing resilience to distribution shifts. 
 On the Cars dataset, our method improves performance by 2–3 percentage points over the second best method. 
 \csm{These results demonstrate that our method not only adheres to strict memory constraints but also effectively retains knowledge across multiple tasks, making it a viable option for real-world applications that prioritize memory efficiency. In contrast, methods like MaxAbs show poor performance, even lagging behind LwF, despite requiring memory proportional to the number of tasks, $\Theta(N)$.}

Next, we assess the effectiveness of various fine-tuning strategies for continual learning using ViT \citep{dosovitskiy2021an} and CLIP \citep{radford2021learning} across two benchmark datasets, CIFAR100 and ImageNet-R. 
The baseline list includes a comprehensive set of established regularization-based continual learning, model-merging, and prompt-based methods.

\begin{table}[!ht]
\centering

\caption{Results for different fine-tuning strategies\label{tab:clip}. Baselines results are taken from \citep{zhou2025learning,marczak2024magmax, DBLP:journals/corr/abs-2403-19137}}
\resizebox{\columnwidth}{!}{\begin{tabular}{c | l | c c | c}
\toprule
Memory & Method    & CIFAR100  & ImageNet-R  & Avg \\
\midrule
$\Theta(1)$ & Zero-shot  & 66.91  & 77.73  & 72.32 \\
N/A        & Joint      & 90.94  & 87.55  & 89.24 \\
\midrule
$\Theta(N)$ & RandMix \citep{marczak2024magmax}   & 77.04  & 81.88  & 79.46 \\
$\Theta(N)$ & MaxAbs  \citep{marczak2024magmax}     & 76.75  & 82.33  & 79.54 \\
$\Theta(N)$ & Avg \cite{wortsman2022model}       & 77.04  & 81.87  & 79.46\\
$\Theta(N)$ & TIES   \citep{yadav2023ties}    & \underline{77.23} & 82.27  & 79.75 \\
\midrule
$\Theta(N)$ & iCaRL \citep{rebuffi2017icarl} & 57.60 & 43.71 & 50.66 \\
$\Theta(N)$ & L2P \citep{wang2022learning}& 70.04 & 69.33 & 69.69 \\
$\Theta(N)$ & DualPrompt \citep{wang2022dualprompt} & 74.31 & 76.41 & 75.36 \\
$\Theta(N)$ & CODA-P \citep{smith2023coda} & 76.4 & 79.5 & 77.95 \\
$\Theta(N)$ & PROOF \citep{zhou2025learning}& 76.55 & 79.7 & 78.13 \\
\midrule
$\Theta(1)$ & LwF     \citep{li2017learning}   & 73.45  & \textbf{82.97} & 78.21 \\
$\Theta(1)$ & EWC    \citep{kirkpatrick2017overcoming}     & 76.24  & 82.42  & 79.33 \\
$\Theta(1)$ & Continual-CLIP\citep{thengane2022clip} & 68.26 &  76.94 & 72.6 \\
$\Theta(1)$ & CoOp \citep{du2022learning} & 70.58 & 78.66 & 74.62 \\
$\Theta(1)$ & MaPLe \citep{khattak2022maple0} & 74.52 & 79.71 & 77.12 \\
$\Theta(1)$ & AttriCLIP \citep{wang2023attriclip} & 68.45 & 76.53 & 72.49 \\
$\Theta(1)$ & CLIP-Adapter \citep{gao2024clip} & 68.32 & 78.1 & 73.21 \\
$\Theta(1)$ & VPT \citep{derakhshani2023bayesian} & 59.33 &  74.0 & 66.66 \\

$\Theta(1)$ & CMM       & \textbf{77.72} & \underline{82.69}  & \textbf{80.21} \\
\bottomrule
\end{tabular}}
\end{table}

As summarized in Table \ref{tab:clip}, our method achieves state-of-the-art performance among $\Theta(1)$ fine-tuning strategies, consistently outperforming or matching more memory-intensive $\Theta(N)$ methods. 
\csm{Specifically, on CIFAR-100, we achieve $77.72\%$ accuracy, surpassing not only baselines such as LwF and EWC but also recent prompt-tuning methods like PROOF by over 1\%. Moreover, on ImageNet-R, our method maintains a high accuracy of $82.69\%$, positioning it as the runner-up for this dataset. These results confirm that our approach narrows the performance gap with $\Theta(N)$ methods, often surpassing them, all while maintaining a constant memory footprint.}

\subsection{Domain-incremental experimental results}

Table \ref{tab:results-DIL} reports accuracy (\%) after the final task on four domain-incremental benchmarks: Office-Home, Office-31, DomainNet, and ImageNet-R. As shown, our approach consistently attains high accuracy across diverse domains, outperforming or closely matching the strongest merging-based baselines. Despite using significantly less memory than  \(\Theta(N)\) approaches, our method achieves a superior average accuracy of $84.49\%$, outperforming or matching specialized continual learning techniques and other merging-based baselines.

\begin{table}[!ht]
\centering
\caption{Our proposed method outperforms other merging-based methods in domain-incremental scenarios and achieves similar results to CL methods.
We report task-agnostic accuracy (\%) after the final task. The best results are in \textbf{bold} and the second best \underline{underlined}.\label{tab:results-DIL}}
\resizebox{\columnwidth}{!}{\begin{tabular}{c|l|cc|cc|c}
\toprule
Memory & Method   & Office-Home & Office-31 & DomainNet & ImageNet-R & Avg \\
\midrule
\(\Theta(N)\) & RandMix \citep{marczak2024magmax}  & \textbf{89.45}     & 93.41     & 64.31     & 82.28            & 82.36 \\
\(\Theta(N)\) & MaxAbs \citep{marczak2024magmax}  & 89.09     & 90.63     & 67.51     & 83.93            & 82.79 \\
\(\Theta(N)\) & Avg \citep{wortsman2022model}      & 85.21     & 89.87     & 64.98     & 82.98            & 80.76 \\
\(\Theta(N)\) & TIES \citep{ties}    & 88.59     & 89.74     & 66.42     & 83.90            & 82.16 \\
\midrule
\(\Theta(1)\) & LwF \citep{li2017learning} & 88.56     & \underline{93.92}     & \textbf{{69.76}}        & \underline{84.78} & \underline{84.25} \\
\(\Theta(1)\) & EWC \citep{kirkpatrick2017overcoming}      & 88.06     & 92.78     & 56.58       & 83.77            & 80.29 \\
\(\Theta(1)\) & CMM     & \underline{89.29}     & \textbf{94.05 }       & \underline{67.27}     & \textbf{{87.35} }           & \textbf{84.49} \\
\bottomrule
\end{tabular}
}
\end{table}
    
From the per-dataset results, we see that our method consistently ranks among the top performers. On Office-31, it achieves $94.05\%$, which is the best result overall, while on ImageNet-R it reaches $87.35\%$, again surpassing all baselines by a clear margin. Even on DomainNet—a challenging dataset—our method’s $67.27\%$ accuracy remains competitive, trailing only LwF by a small gap. These strong individual dataset performances culminate in the highest average accuracy among all methods. It is also worth noting that, LwF is a strong baseline for this DIL experiment, despite its simplicity.

\begin{figure}[!ht]
    \centering
    \vspace*{-2.mm}
     \includegraphics[width=1\columnwidth,]{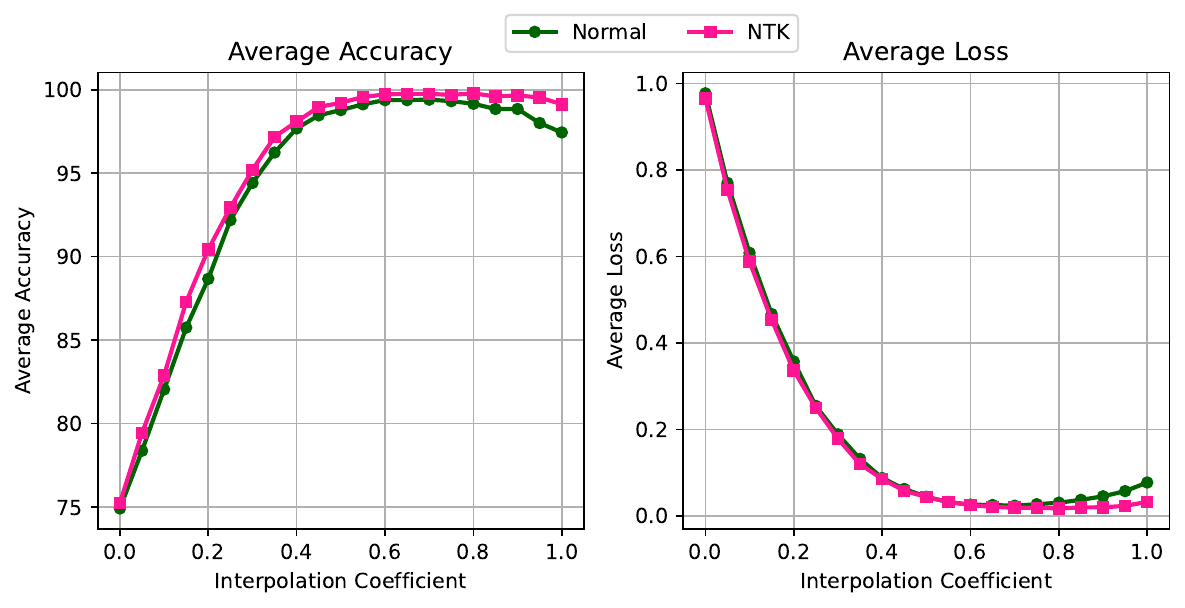}
        \vspace{-3mm}
    \caption{Average accuracy score and loss value when using our merging method in interpolating models after training on first and second tasks using normal training and our proposed linearly fine-tuning method.}
    \label{fig:normal-ntk-fisher}
    \vspace*{-1.mm}    
\end{figure}

\section{Ablation studies}
\csm{To gain a deeper understanding of the contributions made by different components of our proposed method, we conduct a comprehensive ablation study.}
There are three main components of our proposed method, which correspond to three stages in Algorithm \ref{alg:method}. In Figures \ref{fig:normal-ntk-linear} and \ref{fig:normal-ntk-fisher}, we plot the accuracy and cross-entropy loss as a function of the interpolation coefficient. \csm{As shown in these figures, the accuracy curves for NTK-based fine-tuning consistently outperform those of conventional fine-tuning, both for linear weight averaging and our proposed merging method. Similarly, the loss curves follow the same general trend, with NTK fine-tuning consistently yielding lower loss values across the interpolation trajectory. We hypothesize that this improvement can be attributed to the reduction of interference \citep{ortiz2023task} when merging task-specific models.}


\begin{figure}[!ht]
    \centering
    \vspace*{-2.mm}
     \includegraphics[width=1\columnwidth,]{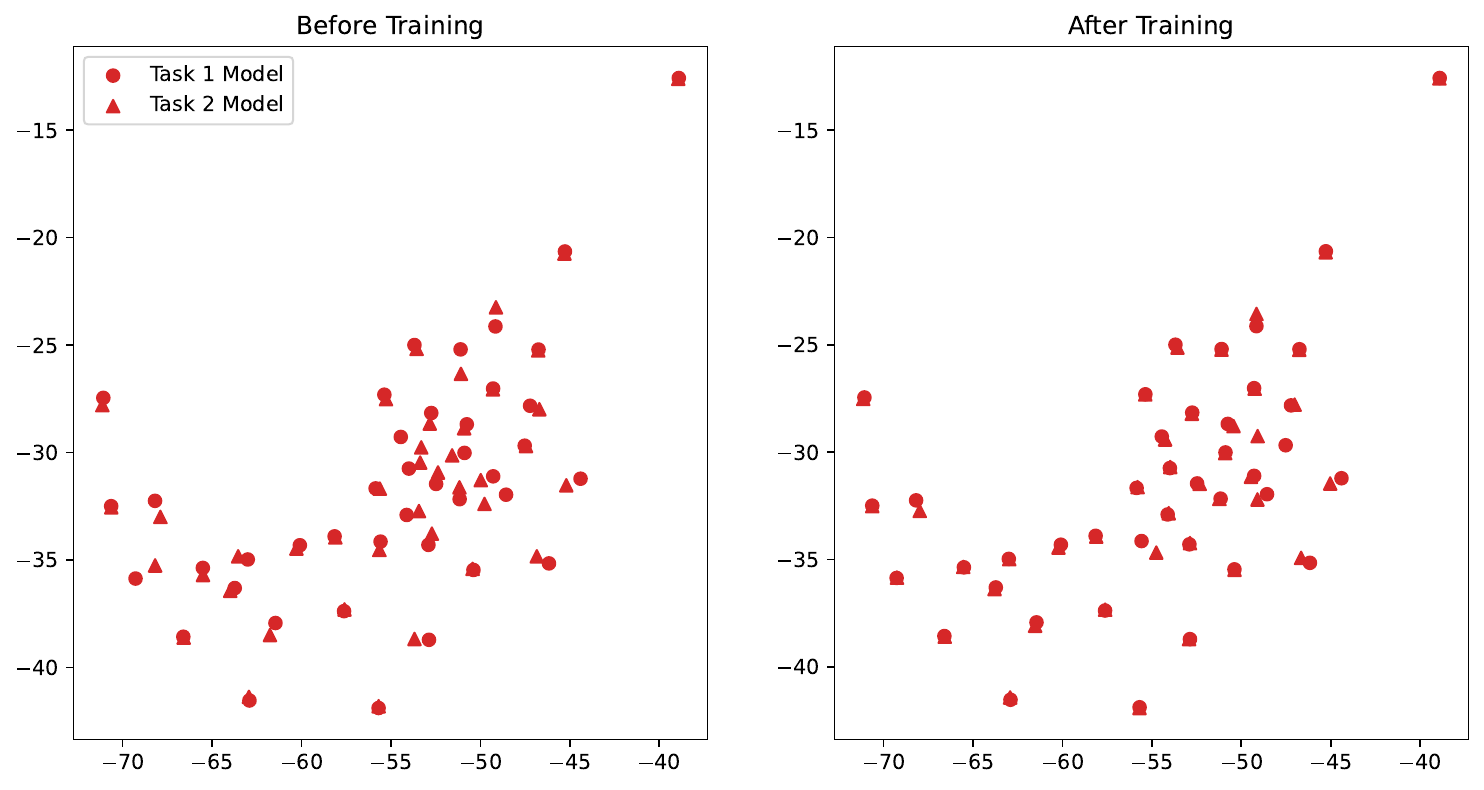}
       \vspace*{-3.mm}
    \caption{t-SNE of samples from the 12-th class using models from the previous task $\circ$ and current task $\triangle$.\label{fig:tsne-12}}
    \vspace*{-1.mm}    
\end{figure}

\begin{figure}[!ht]
    \centering
    \vspace*{-2.mm}
     \includegraphics[width=1\columnwidth,]{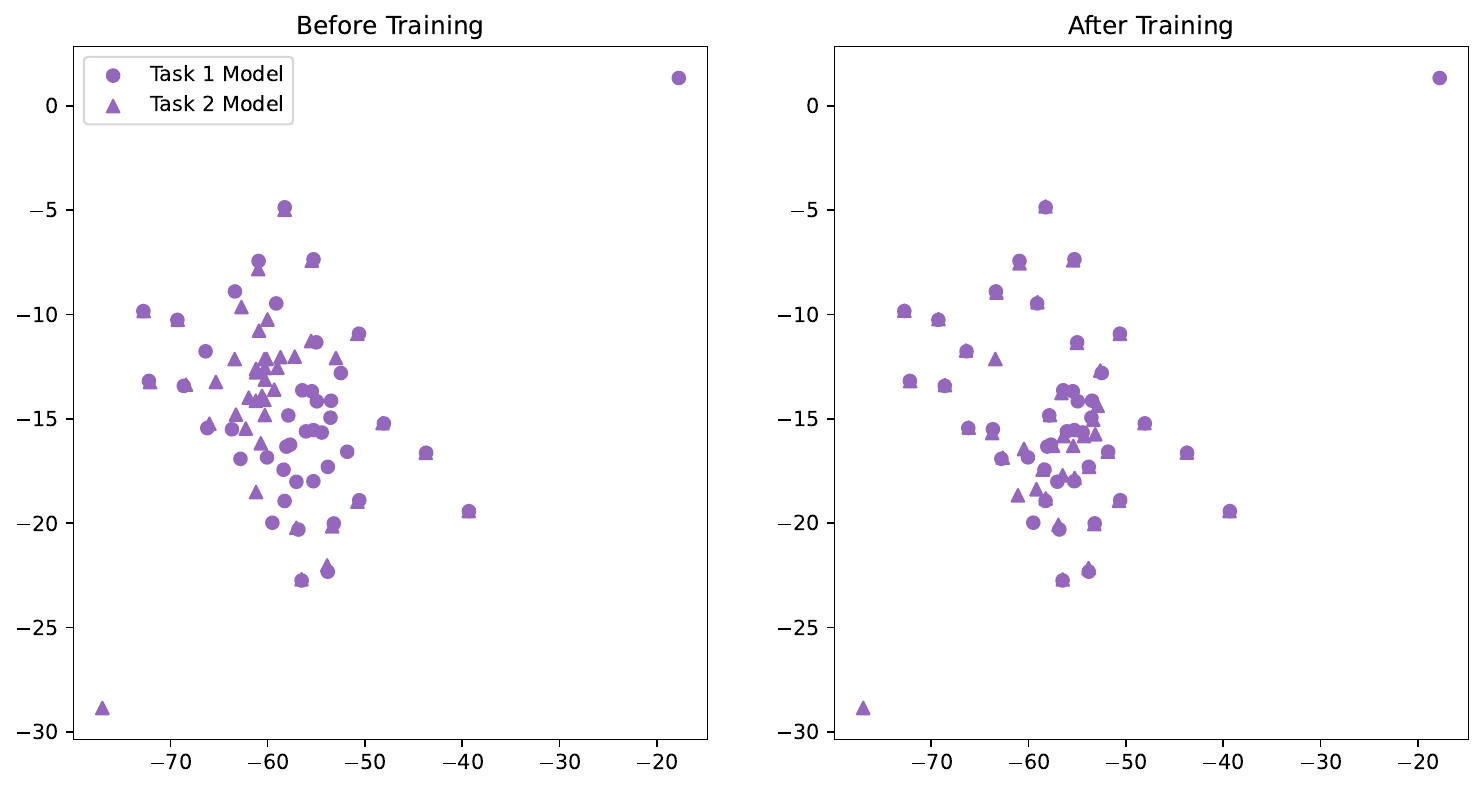}
    \vspace*{-3.mm}
    \caption{t-SNE of samples from the 18-th class using models from the previous task $\circ$ and current task $\triangle$.\label{fig:tsne-18}}
    \label{fig:tsne}
    \vspace*{-1.mm}    
\end{figure}

\begin{figure}[!ht]
    \centering
    \vspace*{-2.5mm}
     \includegraphics[width=1\columnwidth,]{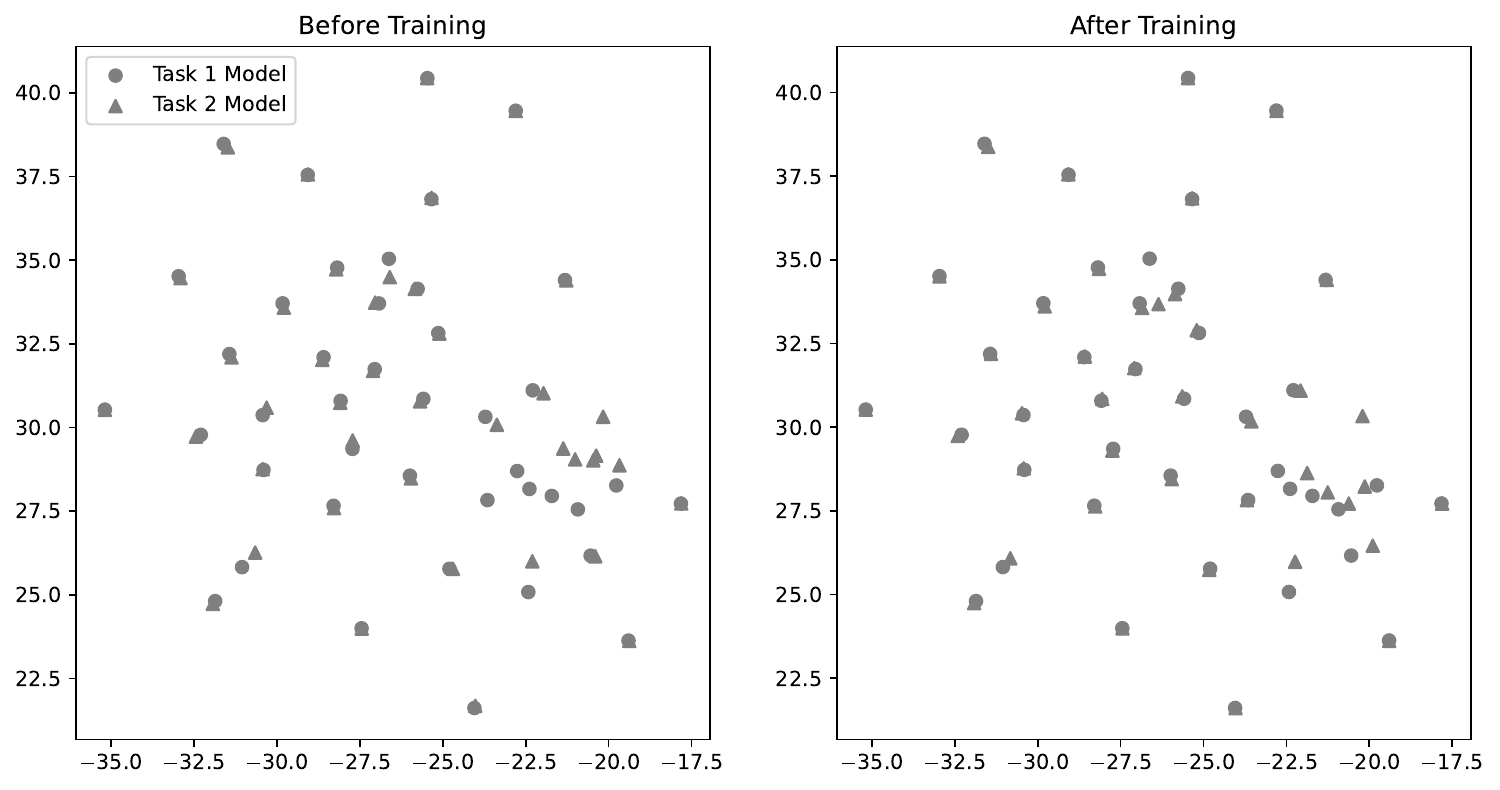}
    \vspace*{-3.5mm}
    \caption{t-SNE of samples from the 29-th class using models from the previous task $\circ$ and current task $\triangle$.\label{fig:tsne-29}}
    \vspace*{-1.5mm}    
\end{figure}

Next, we examine the impact of feature representation refinement on embedding alignment. Figures \ref{fig:tsne-12}, \ref{fig:tsne-18} \csm{and} \ref{fig:tsne-29} \csm{illustrate} the t-SNE of embeddings extracted from three different classes randomly sampled from the Cars dataset. \csm{These visualizations help assess the alignment of the learned feature representations across tasks.} 

\csm{From the left-hand panels, we observe that the feature representation distributions of the merged model (represented by $\triangle$ points) and the model from the previous task (represented by $\circ$ points) exhibit significant misalignment. The merged model’s embeddings are scattered, and there is little alignment with the previous task’s embeddings, suggesting that merging distorts the representations.} In contrast, \csm{note that} the right-hand figures show that these triangle points are brought closer to the circle points following our post-merging procedure. By fine-tuning the lightweight image projection layer, the triangle points shift toward the circle points, confirming the effectiveness of the third stage in our proposed algorithm.

\csm{Table \ref{tab:ablation} presents a detailed breakdown of the impact of each component of our method.}
\csm{As expected, fine-tuning results in a clear performance improvement over zero-shot predictions.}
When merging fine-tuned current and previous models with our FIM approximation, a significant performance boost is observed. 
Next, task-specific weight disentanglement and representation alignment help improve accuracy by reducing task interference and aligning learned representations, respectively.

\begin{table}[!ht]
\caption{Ablative studies of our proposed method on Office-31. Each individual component of our proposed method contributes a distinct performance gain while combining all components yields the highest overall accuracy.\label{tab:ablation}
}
\resizebox{1\columnwidth}{!} {\begin{tabular}{cccc|ccccc}
\toprule
\textbf{FT} & \textbf{Merge} & \textbf{NTK}  & \textbf{Bias} &\textbf{Amazon} & \textbf{Dslr} & \textbf{Webcam}  & \textbf{Avg} \\
\midrule
$\times$ & $\times$ & $\times$ & $\times$ & 80.79 & 83.15 & 83.89 & 82.40 \\
$\checkmark$ & $\times$ & $\times$ & $\times$  & 88.59 & 93.26 & 96.64 & 90.51 \\
$\checkmark$ & $\checkmark$ & $\times$ & $\times$  & 91.53 & 95.28 & 91.95 & 92.91  \\
$\checkmark$ & $\checkmark$ & $\checkmark$ &$\times$  & 91.30 & 96.63 & 97.32 & 93.79  \\
$\checkmark$ & $\checkmark$ & $\checkmark$ & $\checkmark$  & 91.67 & 96.63 & 97.32 & 94.05  \\
\bottomrule
\vspace{-0.2in}
\end{tabular}}
\end{table}

%% file: sec/6_conclusion.tex
\section{Conclusion}
\label{sec:conclusion}
In this work,  we tackle the challenges of applying model-merging techniques within the context of continual learning. 
Drawing on our findings, we propose CMM - a scalable and effective approach that provides a closed-form solution approximating the minimizer of the weighted task loss. 
Our extensive experimental results demonstrate the superiority of our method across various class-incremental and domain-incremental scenarios.

\csm{Despite these successes, there are opportunities for further enhancement, particularly in adapting pre-trained models such as CLIP to novel downstream tasks exhibiting significant covariate shifts that fall outside the scope of CLIP's original training set. The reliance on linear fine-tuning and other parameter-efficient methods, such as adapter and prompt-tuning, may limit the modeling capacity of the original CLIP model. This limitation presents a challenge in effectively adapting to new tasks and highlights potential areas for future research.}

%% file: sec/7_supp.tex
\section{Supplementary for Toward a Holistic Approach to Continual Model Merging}
Due to space constraints, some details were omitted from the main paper. We therefore include baselines' description used in the main paper (Appendix \ref{sec:baseline}) and detailed experimental setup (Appendix \ref{sec:benchmark}) in this supplementary material.
\subsection{Baseline overview}
\label{sec:baseline}
We compare our proposed method with various continual learning approaches, including general methods, model merging techniques, and prompt learning method that can be applied to continual learning.

\begin{itemize}
    \item Learning without Forgetting (LwF) \citep{li2017learning} overcomes forgetting with a distillation loss between the current model and the previously trained one, both applied to the current task's data.
    \item Elastic Weight Consolidation (EWC) \citep{kirkpatrick2017overcoming} regularizes weight changes with a task-importance mask that is accumulatively computed using the diagonal empirical Fisher Information matrix at the end of each task.
    \item iCaRL \citep{rebuffi2017icarl} combines a nearest-mean-of-exemplars classifier with knowledge distillation to retain performance on old classes while learning new ones.
    \item Model Soup (Avg) \citep{wortsman2022model} simply averages task-specific model weights to obtain the merged one.
    \item RandMix \citep{marczak2024magmax} constructs the final weight by randomly choosing one parameter from one of the fine-tuned models.
    \item MaxAbs \citep{marczak2024magmax} constructs the final task vector by selecting the largest magnitude value from all individual task vectors for each entry.
    \item TIES-Merging (TIES) \citep{yadav2023ties} mitigates interference between task vectors during model merging by selectively pruning parameters and determining optimal parameter signs.
    \item Learning to Prompt (L2P) \citep{wang2022learning} introduced the idea of using a shared pool of prompts, from which the top-$k$ most relevant prompts are chosen for each sample during both training and testing. While this method may facilitate knowledge transfer across tasks, it also poses a risk of catastrophic forgetting.
    \item DualPrompt \citep{wang2022dualprompt} addresses the limitations of L2P by attaching auxiliary prompts directly to the pretrained backbone, rather than relying solely on input-level prompting. Additionally, it introduces separate prompt sets for each task, allowing the model to capture more task-specific instructions alongside the task-invariant knowledge drawn from the shared prompt pool.
    \item CODA-Prompt \citep{smith2023coda} uses learnable prompts tailored to each task. Similar to L2P, CODA maintains a pool of prompts and keys, generating the final prompt as a weighted sum based on the cosine similarity between queries and keys. To bypass explicit task prediction at test time, it computes the weighted sum using all prompts, regardless of the task.
    \item CoOp \citep{du2022learning} shares a set of learnable vector tokens for all task classes.
    \item MaPLe \citep{khattak2022maple0} employs two set of learnable prompts for visual and textual encoders, respectively.
    \item CLIP-Adapter \citep{gao2024clip} incorporates lightweight modules over textual and visual branches to enrich respective features of the frozen CLIP model.
    \item Variational Prompt Tuning (VPT) \citep{derakhshani2023bayesian} learns a distribution over input prompt tokens for generalizability, but also faces a trade-off with in-domain performance.

    \item AttriCLIP \citep{wang2023attriclip} is a non-incremental learner designed for CL, built on top of a frozen CLIP model with fixed image and text encoders. It introduces learnable attribute prompts selected from a predefined word bank to enrich category names, enabling classification via visual-textual similarity without updating the classifier head.
    \item PROOF \citep{zhou2025learning} extends LwF to vision-language models by freezing the pretrained image and text encoders and introducing task-specific projection layers for each new task. These projections are trained independently and fused via cross-modal attention, allowing the model to adapt to new tasks while preserving knowledge from previous ones. 
\end{itemize}

\subsection{Domain-incremental benchmarks}
\label{sec:benchmark}
To evaluate our method under domain-incremental setting, we leverage benchmarks from domain adaptation literature: Office-31 \citep{10.5555/1888089.1888106}, Office Home \citep{venkateswara2017deep}, DomainNet \citep{peng2018moment} and ImageNet-R \citep{hendrycks2021many}. For each dataset, tasks are defined as domains. Specifically, Office-31 consists of three different domains, hence is split into 3 tasks. Similarly, Office Home, DomainNet, has 4, 6, and 15 domains, respectively.

%% file: main.bbl
\begin{thebibliography}{10}

\bibitem{achiam2023gpt}
J.~Achiam, S.~Adler, S.~Agarwal, L.~Ahmad, I.~Akkaya, F.~L. Aleman, D.~Almeida, J.~Altenschmidt, S.~Altman, S.~Anadkat, {\em et~al.}, ``Gpt-4 technical report,'' {\em arXiv preprint arXiv:2303.08774}, 2023.

\bibitem{hurst2024gpt}
A.~Hurst, A.~Lerer, A.~P. Goucher, A.~Perelman, A.~Ramesh, A.~Clark, A.~Ostrow, A.~Welihinda, A.~Hayes, A.~Radford, {\em et~al.}, ``Gpt-4o system card,'' {\em arXiv preprint arXiv:2410.21276}, 2024.

\bibitem{touvron2023llama}
H.~Touvron, L.~Martin, K.~Stone, P.~Albert, A.~Almahairi, Y.~Babaei, N.~Bashlykov, S.~Batra, P.~Bhargava, S.~Bhosale, {\em et~al.}, ``Llama 2: Open foundation and fine-tuned chat models,'' {\em arXiv preprint arXiv:2307.09288}, 2023.

\bibitem{grattafiori2024llama}
A.~Grattafiori, A.~Dubey, A.~Jauhri, A.~Pandey, A.~Kadian, A.~Al-Dahle, A.~Letman, A.~Mathur, A.~Schelten, A.~Vaughan, {\em et~al.}, ``The llama 3 herd of models,'' {\em arXiv preprint arXiv:2407.21783}, 2024.

\bibitem{liu2024deepseek}
A.~Liu, B.~Feng, B.~Wang, B.~Wang, B.~Liu, C.~Zhao, C.~Dengr, C.~Ruan, D.~Dai, D.~Guo, {\em et~al.}, ``Deepseek-v2: A strong, economical, and efficient mixture-of-experts language model,'' {\em arXiv preprint arXiv:2405.04434}, 2024.

\bibitem{liu2024deepseek3}
A.~Liu, B.~Feng, B.~Xue, B.~Wang, B.~Wu, C.~Lu, C.~Zhao, C.~Deng, C.~Zhang, C.~Ruan, {\em et~al.}, ``Deepseek-v3 technical report,'' {\em arXiv preprint arXiv:2412.19437}, 2024.

\bibitem{guo2025deepseek}
D.~Guo, D.~Yang, H.~Zhang, J.~Song, R.~Zhang, R.~Xu, Q.~Zhu, S.~Ma, P.~Wang, X.~Bi, {\em et~al.}, ``Deepseek-r1: Incentivizing reasoning capability in llms via reinforcement learning,'' {\em arXiv preprint arXiv:2501.12948}, 2025.

\bibitem{rombach2022high}
R.~Rombach, A.~Blattmann, D.~Lorenz, P.~Esser, and B.~Ommer, ``High-resolution image synthesis with latent diffusion models,'' in {\em Proceedings of the IEEE/CVF conference on computer vision and pattern recognition}, pp.~10684--10695, 2022.

\bibitem{brooks2024video}
T.~Brooks, B.~Peebles, C.~Holmes, W.~DePue, Y.~Guo, L.~Jing, D.~Schnurr, J.~Taylor, T.~Luhman, E.~Luhman, {\em et~al.}, ``Video generation models as world simulators,'' {\em OpenAI Blog}, vol.~1, p.~8, 2024.

\bibitem{li2023deep}
W.~Li, Y.~Peng, M.~Zhang, L.~Ding, H.~Hu, and L.~Shen, ``Deep model fusion: A survey,'' {\em arXiv preprint arXiv:2309.15698}, 2023.

\bibitem{yang2024model}
E.~Yang, L.~Shen, G.~Guo, X.~Wang, X.~Cao, J.~Zhang, and D.~Tao, ``Model merging in llms, mllms, and beyond: Methods, theories, applications and opportunities,'' {\em arXiv preprint arXiv:2408.07666}, 2024.

\bibitem{yang2023adamerging}
E.~Yang, Z.~Wang, L.~Shen, S.~Liu, G.~Guo, X.~Wang, and D.~Tao, ``Adamerging: Adaptive model merging for multi-task learning,'' {\em arXiv preprint arXiv:2310.02575}, 2023.

\bibitem{ahmadian2024mix}
A.~Ahmadian, S.~Goldfarb-Tarrant, B.~Ermis, M.~Fadaee, S.~Hooker, {\em et~al.}, ``Mix data or merge models? optimizing for diverse multi-task learning,'' {\em arXiv preprint arXiv:2410.10801}, 2024.

\bibitem{li2024improving}
M.~Li, Z.~Nie, Y.~Zhang, D.~Long, R.~Zhang, and P.~Xie, ``Improving general text embedding model: Tackling task conflict and data imbalance through model merging,'' {\em arXiv preprint arXiv:2410.15035}, 2024.

\bibitem{li2023revisiting}
Z.~Li, T.~Lin, X.~Shang, and C.~Wu, ``Revisiting weighted aggregation in federated learning with neural networks,'' in {\em International Conference on Machine Learning}, pp.~19767--19788, PMLR, 2023.

\bibitem{jia2024dapperfl}
Y.~Jia, X.~Zhang, H.~Hu, K.-K.~R. Choo, L.~Qi, X.~Xu, A.~Beheshti, and W.~Dou, ``Dapperfl: Domain adaptive federated learning with model fusion pruning for edge devices,'' {\em Advances in Neural Information Processing Systems}, vol.~37, pp.~13099--13123, 2024.

\bibitem{saadati2024dimat}
N.~Saadati, M.~Pham, N.~Saleem, J.~R. Waite, A.~Balu, Z.~Jiang, C.~Hegde, and S.~Sarkar, ``Dimat: Decentralized iterative merging-and-training for deep learning models,'' in {\em Proceedings of the IEEE/CVF Conference on Computer Vision and Pattern Recognition}, pp.~27517--27527, 2024.

\bibitem{chitale2023task}
R.~Chitale, A.~Vaidya, A.~Kane, and A.~Ghotkar, ``Task arithmetic with lora for continual learning,'' {\em arXiv preprint arXiv:2311.02428}, 2023.

\bibitem{araujo2024learning}
V.~Araujo, M.~F. Moens, and T.~Tuytelaars, ``Learning to route for dynamic adapter composition in continual learning with language models,'' in {\em Findings of the Association for Computational Linguistics: EMNLP 2024}, pp.~687--696, 2024.

\bibitem{quarantiello2024adaptive}
L.~Quarantiello, E.~N. Coleman, J.~Hurtado, and V.~Lomonaco, ``Adaptive lora merging for efficient domain incremental learning,'' in {\em Latinx in AI@ NeurIPS 2024}.

\bibitem{goddard2024arcee}
C.~Goddard, S.~Siriwardhana, M.~Ehghaghi, L.~Meyers, V.~Karpukhin, B.~Benedict, M.~McQuade, and J.~Solawetz, ``Arcee’s mergekit: A toolkit for merging large language models,'' in {\em Proceedings of the 2024 Conference on Empirical Methods in Natural Language Processing: Industry Track}, pp.~477--485, 2024.

\bibitem{tang2024fusionbench}
A.~Tang, L.~Shen, Y.~Luo, H.~Hu, B.~Du, and D.~Tao, ``Fusionbench: A comprehensive benchmark of deep model fusion,'' {\em arXiv preprint arXiv:2406.03280}, 2024.

\bibitem{krause20133d}
J.~Krause, M.~Stark, J.~Deng, and L.~Fei-Fei, ``3d object representations for fine-grained categorization,'' in {\em Proceedings of the IEEE international conference on computer vision workshops}, pp.~554--561, 2013.

\bibitem{kirkpatrick2017overcoming}
J.~Kirkpatrick, R.~Pascanu, N.~Rabinowitz, J.~Veness, G.~Desjardins, A.~A. Rusu, K.~Milan, J.~Quan, T.~Ramalho, A.~Grabska-Barwinska, {\em et~al.}, ``Overcoming catastrophic forgetting in neural networks,'' {\em Proceedings of the National Academy of Sciences}, vol.~114, no.~13, pp.~3521--3526, 2017.

\bibitem{rebuffi2017icarl}
S.-A. Rebuffi, A.~Kolesnikov, G.~Sperl, and C.~H. Lampert, ``icarl: Incremental classifier and representation learning,'' in {\em Conference on Computer Vision and Pattern Recognition (CVPR)}, 2017.
\newblock \url{https://arxiv.org/abs/1611.07725}.

\bibitem{li2017learning}
Z.~Li and D.~Hoiem, ``Learning without forgetting,'' {\em IEEE Transactions on Pattern Analysis and Machine Intelligence}, 2017.
\newblock \url{https://arxiv.org/abs/1606.09282}.

\bibitem{rusu2016progressive}
A.~A. Rusu, N.~C. Rabinowitz, G.~Desjardins, H.~Soyer, J.~Kirkpatrick, K.~Kavukcuoglu, R.~Pascanu, and R.~Hadsell, ``Progressive neural networks,'' 2016.
\newblock \url{https://arxiv.org/abs/1606.04671}.

\bibitem{yan2021dynamically}
S.~Yan, J.~Xie, and X.~He, ``Der: Dynamically expandable representation for class incremental learning,'' in {\em Proceedings of the IEEE/CVF Conference on Computer Vision and Pattern Recognition}, pp.~3014--3023, 2021.

\bibitem{buzzega2020dark}
P.~Buzzega, M.~Boschini, A.~Porrello, D.~Abati, and S.~Calderara, ``Dark experience for general continual learning: a strong, simple baseline,'' {\em Advances in neural information processing systems}, vol.~33, pp.~15920--15930, 2020.

\bibitem{shin2017continual}
H.~Shin, J.~K. Lee, J.~Kim, and J.~Kim, ``Continual learning with deep generative replay,'' in {\em Advances in Neural Information Processing Systems (NeurIPS)}, 2017.
\newblock \url{https://arxiv.org/abs/1705.08690}.

\bibitem{wang2022learning}
Z.~Wang, Z.~Zhang, C.-Y. Lee, H.~Zhang, R.~Sun, X.~Ren, G.~Su, V.~Perot, J.~Dy, and T.~Pfister, ``Learning to prompt for continual learning,'' in {\em Proceedings of the IEEE/CVF conference on computer vision and pattern recognition}, pp.~139--149, 2022.

\bibitem{wang2022dualprompt}
Z.~Wang, Z.~Zhang, S.~Ebrahimi, R.~Sun, H.~Zhang, C.-Y. Lee, X.~Ren, G.~Su, V.~Perot, J.~Dy, {\em et~al.}, ``Dualprompt: Complementary prompting for rehearsal-free continual learning,'' in {\em European Conference on Computer Vision}, pp.~631--648, Springer, 2022.

\bibitem{wang2023comprehensive}
L.~Wang, X.~Zhang, H.~Su, and J.~Zhu, ``A comprehensive survey of continual learning: Theory, method and application,'' {\em arXiv preprint arXiv:2302.00487}, 2023.

\bibitem{smith2023coda}
J.~S. Smith, L.~Karlinsky, V.~Gutta, P.~Cascante-Bonilla, D.~Kim, A.~Arbelle, R.~Panda, R.~Feris, and Z.~Kira, ``Coda-prompt: Continual decomposed attention-based prompting for rehearsal-free continual learning,'' in {\em Proceedings of the IEEE/CVF conference on computer vision and pattern recognition}, pp.~11909--11919, 2023.

\bibitem{wortsman2022model}
M.~Wortsman, G.~Ilharco, S.~Y. Gadre, R.~Roelofs, R.~Gontijo-Lopes, A.~S. Morcos, H.~Namkoong, A.~Farhadi, Y.~Carmon, S.~Kornblith, {\em et~al.}, ``Model soups: averaging weights of multiple fine-tuned models improves accuracy without increasing inference time,'' in {\em International Conference on Machine Learning (ICML)}, 2022.
\newblock \url{https://arxiv.org/abs/2203.05482}.

\bibitem{ilharco2023task}
G.~Ilharco, M.~T. Ribeiro, M.~Wortsman, S.~Gururangan, L.~Schmidt, H.~Hajishirzi, and A.~Farhadi, ``Editing models with task arithmetic,'' in {\em International Conference on Learning Representations (ICLR)}, 2023.
\newblock \url{https://arxiv.org/abs/2110.08207}.

\bibitem{matena2022merging}
M.~S. Matena and C.~A. Raffel, ``Merging models with fisher-weighted averaging,'' {\em Advances in Neural Information Processing Systems}, vol.~35, pp.~17703--17716, 2022.

\bibitem{jin2023dataless}
X.~Jin, X.~Ren, D.~Preotiuc-Pietro, and P.~Cheng, ``Dataless knowledge fusion by merging weights of language models,'' in {\em The Eleventh International Conference on Learning Representations}, 2023.

\bibitem{ainsworth2022git}
S.~K. Ainsworth, J.~Hayase, and S.~Srinivasa, ``Git re-basin: Merging models modulo permutation symmetries,'' in {\em International Conference on Learning Representations (ICLR)}, 2023.
\newblock \url{https://arxiv.org/abs/2209.04836}.

\bibitem{yadav2023ties}
P.~Yadav, D.~Tam, L.~Choshen, C.~A. Raffel, and M.~Bansal, ``Ties-merging: Resolving interference when merging models,'' {\em Advances in Neural Information Processing Systems}, vol.~36, pp.~7093--7115, 2023.

\bibitem{dhawan2024leveraging}
N.~Dhawan, N.~E. Mitchell, Z.~Charles, Z.~Garrett, and G.~K. Dziugaite, ``Leveraging function space aggregation for federated learning at scale,'' {\em Transactions on Machine Learning Research}, 2024.
\newblock Expert Certification.

\bibitem{marczak2024magmax}
D.~Marczak, B.~Twardowski, T.~Trzci{\'n}ski, and S.~Cygert, ``Magmax: Leveraging model merging for seamless continual learning,'' in {\em European Conference on Computer Vision}, pp.~379--395, Springer, 2024.

\bibitem{marouf2024weightedensemblemodelsstrong}
I.~E. Marouf, S.~Roy, E.~Tartaglione, and S.~Lathuilière, ``Weighted ensemble models are strong continual learners,'' 2024.

\bibitem{kingma2014adam}
D.~P. Kingma, ``Adam: A method for stochastic optimization,'' {\em arXiv preprint arXiv:1412.6980}, 2014.

\bibitem{loshchilov2018decoupled}
I.~Loshchilov and F.~Hutter, ``Decoupled weight decay regularization,'' in {\em International Conference on Learning Representations (ICLR)}, 2019.

\bibitem{radford2021learning}
A.~Radford, J.~W. Kim, C.~Hallacy, A.~Ramesh, G.~Goh, S.~Agarwal, G.~Sastry, A.~Askell, P.~Mishkin, J.~Clark, {\em et~al.}, ``Learning transferable visual models from natural language supervision,'' in {\em International Conference on Machine Learning}, pp.~8748--8763, PMLR, 2021.

\bibitem{loshchilov2017decoupled}
I.~Loshchilov, ``Decoupled weight decay regularization,'' {\em arXiv preprint arXiv:1711.05101}, 2017.

\bibitem{amari1998natural}
S.-I. Amari, ``Natural gradient works efficiently in learning,'' {\em Neural computation}, vol.~10, no.~2, pp.~251--276, 1998.

\bibitem{eschenhagen2023kronecker}
R.~Eschenhagen, A.~Immer, R.~Turner, F.~Schneider, and P.~Hennig, ``Kronecker-factored approximate curvature for modern neural network architectures,'' {\em Advances in Neural Information Processing Systems}, vol.~36, pp.~33624--33655, 2023.

\bibitem{hwang2024fadam}
D.~Hwang, ``Fadam: Adam is a natural gradient optimizer using diagonal empirical fisher information,'' {\em arXiv preprint arXiv:2405.12807}, 2024.

\bibitem{jacot2018neural}
A.~Jacot, F.~Gabriel, and C.~Hongler, ``Neural tangent kernel: Convergence and generalization in neural networks,'' in {\em Advances in Neural Information Processing Systems (NeurIPS)}, 2018.
\newblock \url{https://proceedings.neurips.cc/paper_files/paper/2018/file/5a4be1fa34e62bb8a6ec6b91d2462f5a-Paper.pdf}.

\bibitem{arora2019exact}
S.~Arora, S.~S. Du, W.~Hu, Z.~Li, R.~Salakhutdinov, and R.~Wang, ``On exact computation with an infinitely wide neural net,'' in {\em Advances in Neural Information Processing Systems (NeurIPS)}, 2019.
\newblock \url{https://proceedings.neurips.cc/paper/2019/hash/dbc4d84bfcfe2284ba11beffb853a8c4-Abstract.html}.

\bibitem{lee2019wide}
J.~Lee, L.~Xiao, S.~Schoenholz, Y.~Bahri, R.~Novak, J.~Sohl-Dickstein, and J.~Pennington, ``Wide neural networks of any depth evolve as linear models under gradient descent,'' in {\em Advances in Neural Information Processing Systems (NeurIPS)}, 2019.
\newblock \url{https://proceedings.neurips.cc/paper_files/paper/2019/file/0d1a9651497a38d8b1c3871c84528bd4-Paper.pdf}.

\bibitem{malladi2022kernel}
S.~Malladi, A.~Wettig, D.~Yu, D.~Chen, and S.~Arora, ``A kernel-based view of language model fine-tuning,'' 2022.
\newblock \url{https://arxiv.org/abs/2210.05643}.

\bibitem{ortizjimenez2021linear}
G.~Ortiz{-}Jim{\'{e}}nez, A.~Modas, S.~Moosavi{-}Dezfooli, and P.~Frossard, ``What can linearized neural networks actually say about generalization?,'' in {\em Advances in Neural Information Processing Systems (NeurIPS)}, 2021.
\newblock \url{https://proceedings.neurips.cc/paper_files/paper/2021/file/4b5deb9a14d66ab0acc3b8a2360cde7c-Paper.pdf}.

\bibitem{zancato2020predicting}
L.~Zancato, A.~Achille, A.~Ravichandran, R.~Bhotika, and S.~Soatto, ``Predicting training time without training,'' in {\em Advances in Neural Information Processing Systems (NeurIPS)}, 2020.
\newblock \url{https://proceedings.neurips.cc/paper/2020/hash/440e7c3eb9bbcd4c33c3535354a51605-Abstract.html}.

\bibitem{deshpande2021linearized}
A.~Deshpande, A.~Achille, A.~Ravichandran, H.~Li, L.~Zancato, C.~C. Fowlkes, R.~Bhotika, S.~Soatto, and P.~Perona, ``A linearized framework and a new benchmark for model selection for fine-tuning,'' 2021.
\newblock \url{https://arxiv.org/abs/2102.00084}.

\bibitem{yuce2022inrs}
G.~Y{\"{u}}ce, G.~Ortiz{-}Jim{\'{e}}nez, B.~Besbinar, and P.~Frossard, ``A structured dictionary perspective on implicit neural representations,'' in {\em {IEEE} Conference on Computer Vision and Pattern Recognition ({CVPR})}, 2022.
\newblock \url{https://doi.org/10.1109/CVPR52688.2022.01863}.

\bibitem{kunstner2019limitations}
F.~Kunstner, P.~Hennig, and L.~Balles, ``Limitations of the empirical fisher approximation for natural gradient descent,'' {\em Advances in neural information processing systems}, vol.~32, 2019.

\bibitem{ortiz2023task}
G.~Ortiz-Jimenez, A.~Favero, and P.~Frossard, ``Task arithmetic in the tangent space: Improved editing of pre-trained models,'' {\em Advances in Neural Information Processing Systems}, vol.~36, pp.~66727--66754, 2023.

\bibitem{jha2025clap4clip}
S.~Jha, D.~Gong, and L.~Yao, ``Clap4clip: Continual learning with probabilistic finetuning for vision-language models,'' {\em Advances in neural information processing systems}, vol.~37, pp.~129146--129186, 2025.

\bibitem{ties}
P.~Yadav, D.~Tam, L.~Choshen, C.~Raffel, and M.~Bansal, ``Resolving interference when merging models,'' in {\em Advances in Neural Information Processing Systems (NeurIPS)}, 2023.
\newblock \url{https://arxiv.org/abs/2306.01708}.

\bibitem{krizhevsky2009learning}
A.~Krizhevsky, G.~Hinton, {\em et~al.}, ``Learning multiple layers of features from tiny images,'' 2009.
\newblock \url{https://www.cs.toronto.edu/~kriz/learning-features-2009-TR.pdf}.

\bibitem{hendrycks2021many}
D.~Hendrycks, S.~Basart, N.~Mu, S.~Kadavath, F.~Wang, E.~Dorundo, R.~Desai, T.~Zhu, S.~Parajuli, M.~Guo, {\em et~al.}, ``The many faces of robustness: A critical analysis of out-of-distribution generalization,'' in {\em Proceedings of the IEEE/CVF international conference on computer vision}, pp.~8340--8349, 2021.

\bibitem{wah2011caltech}
C.~Wah, S.~Branson, P.~Welinder, P.~Perona, and S.~Belongie, ``The caltech-ucsd birds-200-2011 dataset,'' 2011.

\bibitem{10.5555/1888089.1888106}
K.~Saenko, B.~Kulis, M.~Fritz, and T.~Darrell, ``Adapting visual category models to new domains,'' in {\em Proceedings of the 11th European Conference on Computer Vision: Part IV}, ECCV'10, (Berlin, Heidelberg), p.~213–226, Springer-Verlag, 2010.

\bibitem{venkateswara2017deep}
H.~Venkateswara, J.~Eusebio, S.~Chakraborty, and S.~Panchanathan, ``Deep hashing network for unsupervised domain adaptation,'' {\em arXiv preprint arXiv: 1706.07522}, 2017.

\bibitem{peng2018moment}
X.~Peng, Q.~Bai, X.~Xia, Z.~Huang, K.~Saenko, and B.~Wang, ``Moment matching for multi-source domain adaptation,'' {\em IEEE International Conference on Computer Vision}, 2018.

\bibitem{wang2023attriclip}
R.~Wang, X.~Duan, G.~Kang, J.~Liu, S.~Lin, S.~Xu, J.~L{\"u}, and B.~Zhang, ``Attriclip: A non-incremental learner for incremental knowledge learning,'' in {\em Proceedings of the IEEE/CVF Conference on Computer Vision and Pattern Recognition}, pp.~3654--3663, 2023.

\bibitem{dosovitskiy2020image}
A.~Dosovitskiy, L.~Beyer, A.~Kolesnikov, D.~Weissenborn, X.~Zhai, T.~Unterthiner, M.~Dehghani, M.~Minderer, G.~Heigold, S.~Gelly, {\em et~al.}, ``An image is worth 16x16 words: Transformers for image recognition at scale,'' {\em arXiv preprint arXiv:2010.11929}, 2020.

\bibitem{ilharco2022patching}
G.~Ilharco, M.~Wortsman, S.~Y. Gadre, S.~Song, H.~Hajishirzi, S.~Kornblith, A.~Farhadi, and L.~Schmidt, ``Patching open-vocabulary models by interpolating weights,'' in {\em Advances in Neural Information Processing Systems (NeurIPS)}, 2022.
\newblock \url{https://arXiv.org/abs/2208.05592}.

\bibitem{ilharco_gabriel_2021_5143773}
G.~Ilharco, M.~Wortsman, R.~Wightman, C.~Gordon, N.~Carlini, R.~Taori, A.~Dave, V.~Shankar, H.~Namkoong, J.~Miller, H.~Hajishirzi, A.~Farhadi, and L.~Schmidt, ``Openclip,'' 2021.
\newblock \url{https://github.com/mlfoundations/open_clip}.

\bibitem{hu2019overcoming}
W.~Hu, Z.~Lin, B.~Liu, C.~Tao, Z.~T. Tao, D.~Zhao, J.~Ma, and R.~Yan, ``Overcoming catastrophic forgetting for continual learning via model adaptation,'' in {\em International Conference on Learning Representations}, 2019.

\bibitem{dosovitskiy2021an}
A.~Dosovitskiy, L.~Beyer, A.~Kolesnikov, D.~Weissenborn, X.~Zhai, T.~Unterthiner, M.~Dehghani, M.~Minderer, G.~Heigold, S.~Gelly, J.~Uszkoreit, and N.~Houlsby, ``An image is worth 16x16 words: Transformers for image recognition at scale,'' in {\em International Conference on Learning Representations (ICLR)}, 2021.
\newblock \url{https://openreview.net/forum?id=YicbFdNTTy}.

\bibitem{zhou2025learning}
D.-W. Zhou, Y.~Zhang, Y.~Wang, J.~Ning, H.-J. Ye, D.-C. Zhan, and Z.~Liu, ``Learning without forgetting for vision-language models,'' {\em IEEE Transactions on Pattern Analysis and Machine Intelligence}, 2025.

\bibitem{DBLP:journals/corr/abs-2403-19137}
S.~Jha, D.~Gong, and L.~Yao, ``Clap4clip: Continual learning with probabilistic finetuning for vision-language models,'' {\em CoRR}, vol.~abs/2403.19137, 2024.

\bibitem{thengane2022clip}
V.~Thengane, S.~Khan, M.~Hayat, and F.~Khan, ``Clip model is an efficient continual learner,'' {\em arXiv preprint arXiv:2210.03114}, 2022.

\bibitem{du2022learning}
Y.~Du, F.~Wei, Z.~Zhang, M.~Shi, Y.~Gao, and G.~Li, ``Learning to prompt for open-vocabulary object detection with vision-language model,'' in {\em Proceedings of the IEEE/CVF Conference on Computer Vision and Pattern Recognition}, pp.~14084--14093, 2022.

\bibitem{khattak2022maple0}
M.~U. Khattak, H.~Rasheed, M.~Maaz, S.~H. Khan, and F.~Khan, ``Maple: Multi-modal prompt learning,'' {\em Computer Vision and Pattern Recognition}, 2022.

\bibitem{gao2024clip}
P.~Gao, S.~Geng, R.~Zhang, T.~Ma, R.~Fang, Y.~Zhang, H.~Li, and Y.~Qiao, ``Clip-adapter: Better vision-language models with feature adapters,'' {\em International Journal of Computer Vision}, vol.~132, no.~2, pp.~581--595, 2024.

\bibitem{derakhshani2023bayesian}
M.~M. Derakhshani, E.~Sanchez, A.~Bulat, V.~G.~T. da~Costa, C.~G. Snoek, G.~Tzimiropoulos, and B.~Martinez, ``Bayesian prompt learning for image-language model generalization,'' in {\em Proceedings of the IEEE/CVF International Conference on Computer Vision}, pp.~15237--15246, 2023.

\end{thebibliography}
